%% file: article.tex
\documentclass[runningheads]{llncs}
\usepackage{graphicx}

\usepackage{tikz}
\usepackage{comment}
\usepackage{amsmath,amssymb} 
\usepackage{color}
\usepackage{todonotes}
\usepackage{caption}
\usepackage{subcaption}
\usepackage{mathrsfs}
\usepackage{fancyhdr}

\fancypagestyle{firstpagestyle}
{
  
  \fancyhf{}
}

\newcommand\bodytextwo{3DBodyTex.v2}
\newcommand\methodname{3DBooSTeR}

\DeclareMathOperator*{\argmin}{arg\,min}

\usepackage{xspace}
\makeatletter
\DeclareRobustCommand\onedot{\futurelet\@let@token\@onedot}
\def\@onedot{\ifx\@let@token.\else.\null\fi\xspace}
\def\eg{\emph{e.g}\onedot}

\def\ie{\emph{i.e}\onedot}

\begin{document}
\pagestyle{headings}
\mainmatter
\def\ECCVSubNumber{8}  

\title{\methodname{}:\\ 3D Body Shape and Texture Recovery}

\author{
Alexandre Saint\inst{1} \and
Anis Kacem\inst{1} \and
Kseniya Cherenkova\inst{1,2} \and
Djamila Aouada\inst{1}
}

\authorrunning{A. Saint et al.}
%

\institute{
SnT, University of Luxembourg\\
\email{\{alexandre.saint,anis.kacem,djamila.aouada\}@uni.lu}
\and
Artec3D \\
\email{kcherenkova@artec-group.com}
}
\maketitle

\thispagestyle{firstpagestyle}

\input{abstract}
\input{Introduction}
\input{ProblemStatement}
\input{ProposedApproach}
\input{Experiments}
\input{Conclusion}
\input{acknowledgements}

\bibliographystyle{splncs04}
\bibliography{egbib}
\end{document}

%% file: abstract.tex
\begin{abstract}


We propose 3DBooSTeR,
a novel method to recover a textured 3D body mesh from a textured partial 3D scan.
With the advent of virtual and augmented reality, there is a demand for
creating realistic and high-fidelity digital 3D human representations.
However, 3D scanning systems can only capture the 3D human body shape up to some
level of defects due to its complexity, including
occlusion between body parts,
varying levels of details,
shape deformations
and the articulated skeleton.
Textured 3D mesh completion is thus important to enhance 3D acquisitions.
The proposed approach decouples the shape and texture completion into
two sequential tasks.
The shape is recovered by an encoder-decoder network deforming a template
body mesh.
The texture is subsequently obtained by projecting the partial texture onto
the template mesh before inpainting the corresponding texture map with a novel
approach.
The approach is validated on the 3DBodyTex.v2 dataset.

\keywords{
  3D shape completion
  \and Human body shape
  \and Point cloud
  \and Texture
  \and Inpainting
}
\end{abstract}

%% file: Introduction.tex
\begin{figure}[h!]
    \centering
    \includegraphics[width=0.9\linewidth]{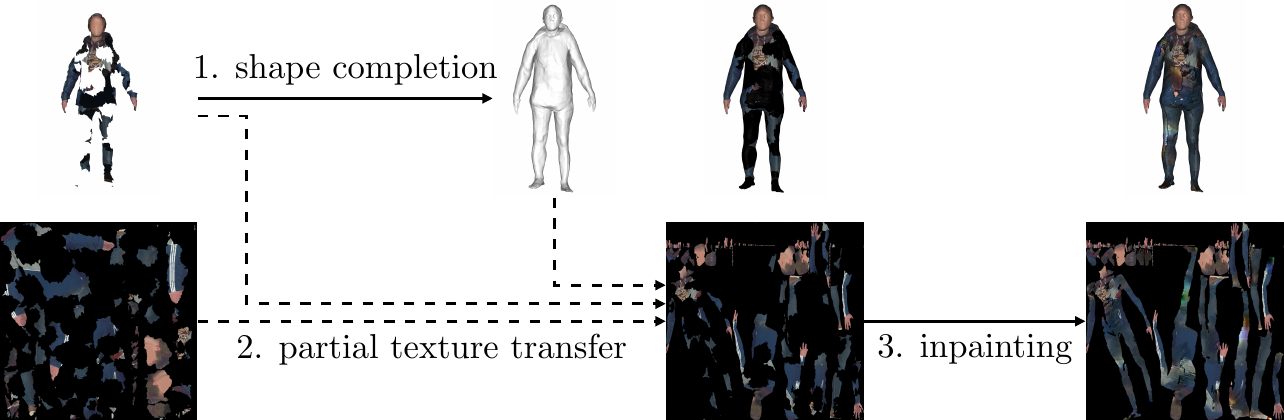}
    \caption{
    Overview of the proposed approach for completing a partial textured 3D body mesh.
    1) The complete shape is estimated.
    2) The partial texture is transferred onto the estimated shape.
    3) The corresponding texture image is inpainted.
    }
    \label{fig:approach_highlevel}
\end{figure}

\section{Introduction}

The completion of a partial textured 3D body shape is key to enable the digital
representation of 3D body shape with realistic details, in a reasonable time
and in an automated way.
This is required in applications such as virtual reality.
Capturing the 3D human shape and texture of the human body is complex due
to the occlusion of body parts,
the complex shape of the clothing wrinkles,
the variation of the pose in time, etc.
Some photogrammetric scanner systems, on the one hand, use a large array of cameras
to cover the body shape from all possible angles.
These systems are able to capture textured 3D shape sequences at high frame rates (120kHz).
However, they still suffer from occlusion and cannot represent fine details,
such as the fingers and the ears.
Hand-held scanners, on the other hand, can be brought to resolve fine details and cover all view angles, but they require a static target.
For an unconstrained usage, it is then more desirable to acquire a partial shape and come with an effective completion solution that recovers the missing data.
Aware of this emergent need, some state-of-art works have tried to recover the missing information provided by the scanning devices.
Most of them consider the problems of shape and texture completion independently.
Accordingly, the relevant literature is presented separately below.

\paragraph{Shape completion}
A simple approach to shape completion is the hole-filling algorithm (\eg Davis~et~al.~\cite{davis2002filling}), in which the missing shape regions are filled with a surface patch joining the boundaries of the available surface.
This approach is limited to relatively small holes with respect to the surrounding surface and to relatively smooth regions.

An approach to regularise the shape completion is to rely on a template shape that is deformed to match the input partial shape.
Szeliski~et~al.~\cite{szeliski1996matching} deform a simple convex shape to represent anatomical body parts such as the head.
Anguelov~et~al.~\cite{anguelov2005scape} learn a parametric model of the human body pose and shape to regularise the completion of a full 3D body shape while handling the large deformations caused by variation in pose.
This sort of methods usually requires manual intialisation,
as shown by Saint~et~al.~\cite{saint20183dbodytex,saint2019bodyfitr},
who propose fully-automatic body model fitting approaches
by exploiting the colour or texture information in human scans,
as available in 3DBodyTex~\cite{saint20183dbodytex}.
This allows recovering shapes with relatively large proportions of missing data.
However, body models smooth out the shape details and are limited to the body shape without clothing.
In these works, the completion of the texture is not considered,
even though the texture might be used to regularise the shape completion~\cite{saint20183dbodytex,saint2019bodyfitr}.

Some works use volumetric convolutions to complete partial 3D shapes~\cite{wu20153d,dai2017shape,han2017high}.
The achievable resolution is limited due to the high computational complexity of 3D convolutions.
Moreover, this category of approaches works well on relatively rigid shapes~\cite{wu20153d,dai2017shape}
(\eg objects of the same class)
but less well on deformable shapes~\cite{litany2018deformable} (\eg shape of the human body or of animals).
Chibane~et~al.~\cite{chibane2020implicit} represent the 3D body shape with an implicit function.
The implicit function is approximated by a deep neural network and learned from a dataset of example 3D body shapes.
This method allows completing partial shapes but does not consider the texture.
It accepts different input 3D shape representations (\eg point cloud, mesh, voxel grid),
however, the output surface must be recovered from the implicit function with a post-processing step, such as the marching cube algorithm~\cite{lorensen1987marching}.

Several works learn the space of body shape deformations
using deep learning~\cite{groueix20183d,li2019lbs,litany2018deformable,ma2020learning}.
These models use encoder-decoder architectures.
An input shape is encoded into a latent representation by the encoder.
The decoder then deforms a base body mesh from a canonical pose and shape to a specific pose and identity, 
using the intermediate representation as input.
The deformations are performed with mesh convolutions
(\eg PointNet~\cite{qi2017pointnet} or FeaStNet~\cite{verma2018feastnet}).
The parameters of the network are learned from a dataset of example 3D body shapes.
Some works target only the body shape with minimal close-fitting
clothing~\cite{groueix20183d,li2019lbs,litany2018deformable},
which is locally smooth and regular. 
Other works target the body shape with casual clothing~\cite{ma2020learning},
which contains irregular local variations, such as wrinkles, due to factors including the cut and the fabric.
The completion of texture information is not tackled in these works.

\paragraph{Texture completion}
Deng~et~al.~\cite{deng2018uv} recover the 3D shape and the colour information of a face
from a non-frontal 2D view.
Shape and texture completion are decoupled.
A 3D morphable model (3DMM) is first fitted to the image.
Then, the available colour information is projected onto the UV map of the template mesh.
Finally, the UV map is inpainted to recover the missing colour information.
The fitting of a body is more complex due to the pose variation, larger and more non-linear
than the variation in the expression.
The UV map for the face is a single chart.
For a body model, it is a set of multiple charts representing different body regions.

In the context of partial shape and texture completion, we present our approach to solve the SHARP challenge~\cite{Sharp2020} on recovering large regions of partial textured body meshes.
This challenge provides \bodytextwo{}, a dataset with thousands of textured body meshes.
The training set and the validation set contain the ground-truth textured body meshes.
The evaluation set contains only partial meshes.
The goal is to estimate the complete shapes in the evaluation set. 

Our contribution, sketched in Fig.~\ref{fig:approach_highlevel},
is a method, named \methodname{}, to recover a 3D body mesh with a corresponding high-resolution texture
from a textured partial 3D body scan.
The tasks of shape and texture completion are decoupled into a sequential pipeline.
The shape completion method is a data-driven mesh deformation deep learning network
(based on Groueix~et~al.~\cite{groueix20183d}) that produces an output mesh of fixed topology.
The texture completion is reduced to an inpainting task of the texture image of the reconstructed mesh.
An novel inpainting method (based on Liu~et~al.~\cite{liu2018partialinpainting})
is proposed.
It is specifically designed to handle the inpainting of a texture image with robustness to
irregularly-shaped incomplete regions
and irregularly-shaped background regions that must be ignored.

The rest of the paper is organised as follows:
Section~\ref{sect:problem_statement} introduces the problem and the notations used in the paper.
Section~\ref{sect:approach} presents the proposed approach for shape and texture completion.
Experimental results are reported in Section~\ref{sect:experiments}.
In Section~\ref{sect:conclusion}, we conclude the paper.

%% file: ProblemStatement.tex
\section{Problem Statement}
\label{sect:problem_statement}

A textured body shape is denoted $\mathcal{X} = (\mathcal{S},\mathit{T})$,
where $\mathcal{S}=(\mathbf{V},\mathbf{F},\pi)$ is a body mesh
with $n_v$ vertices stacked in a matrix $\mathbf{V} \in \mathbb{R}^{n_v\times 3}$,
$n_f$ triangular faces encoded in
$\mathbf{F} \in \mathbb{N}^{n_f\times 3}$
as triplets of vertex indices,
and a 2D parametrisation, $\pi$,
defining a mapping of the faces between the 3D shape, $\mathcal{S}$,
and the 2D texture image, $\mathit{T}$.

Given a partial textured body shape $\mathcal{X}_p = (\mathcal{S}_p,\mathit{T}_p)$,
with, $\mathcal{S}_p$, the partial mesh,
and, $\mathit{T}_p$, the partial texture,
we aim to predict a complete textured body mesh
$\hat{\mathcal{X}} = (\hat{\mathcal{S}},\hat{\mathit{T}})$
that approximates well the ground truth
$\mathcal{X} = (\mathcal{S},\mathit{T})$.
Moreover, the estimation $\hat{\mathcal{X}}$ should preserve
the partially provided texture and shape information as much as possible.

\begin{figure}
\centering
\begin{subfigure}{.2\textwidth}
  \centering
  \includegraphics[height=\linewidth]{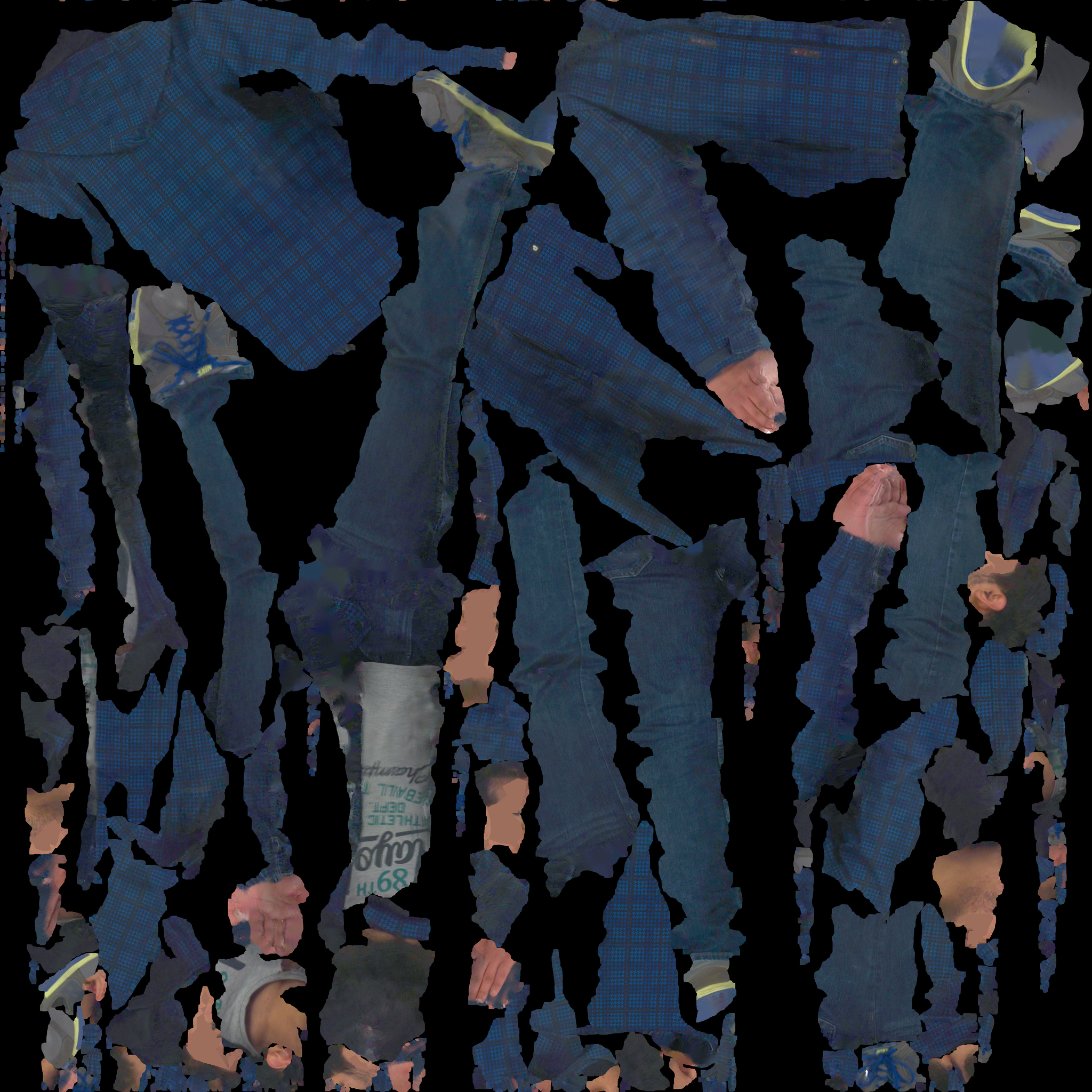}
  \caption{
  A complete texture atlas.
  \\ 
  }
  \label{fig:texture_atlas_complete}
\end{subfigure}%
\hfill
\begin{subfigure}{.2\textwidth}
  \centering
  \includegraphics[height=\linewidth]{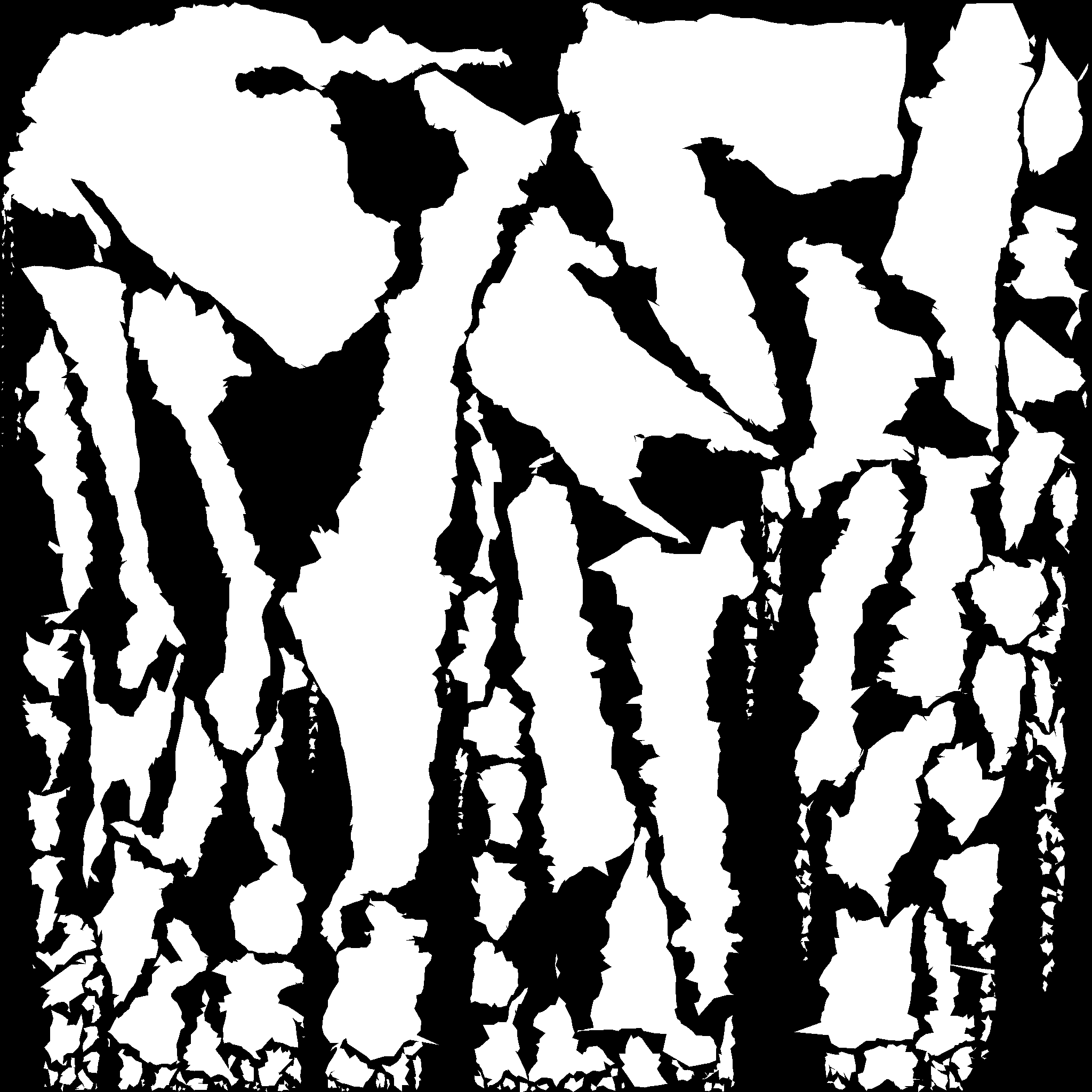}
  \caption{Background mask of complete texture.}
  \label{fig:texture_atlas_complete_mask_background}
\end{subfigure}%
\hfill
\begin{subfigure}{.2\textwidth}
  \centering
  \includegraphics[height=\linewidth]{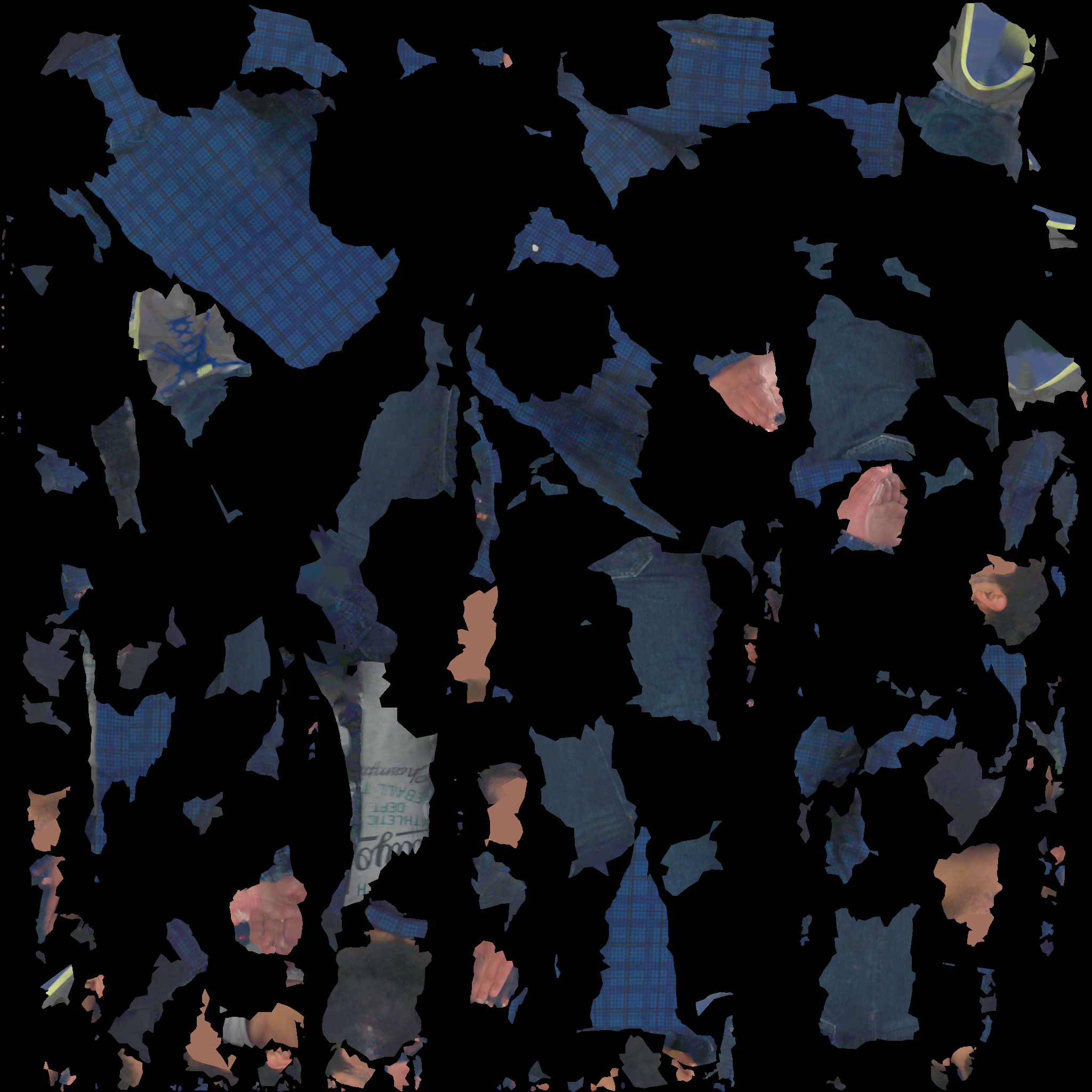}
  \caption{
  A partial texture atlas.
  \\ 
  }
  \label{fig:texture_atlas_partial}
\end{subfigure}
\hfill
\begin{subfigure}{.2\textwidth}
  \centering
  \includegraphics[height=\linewidth]{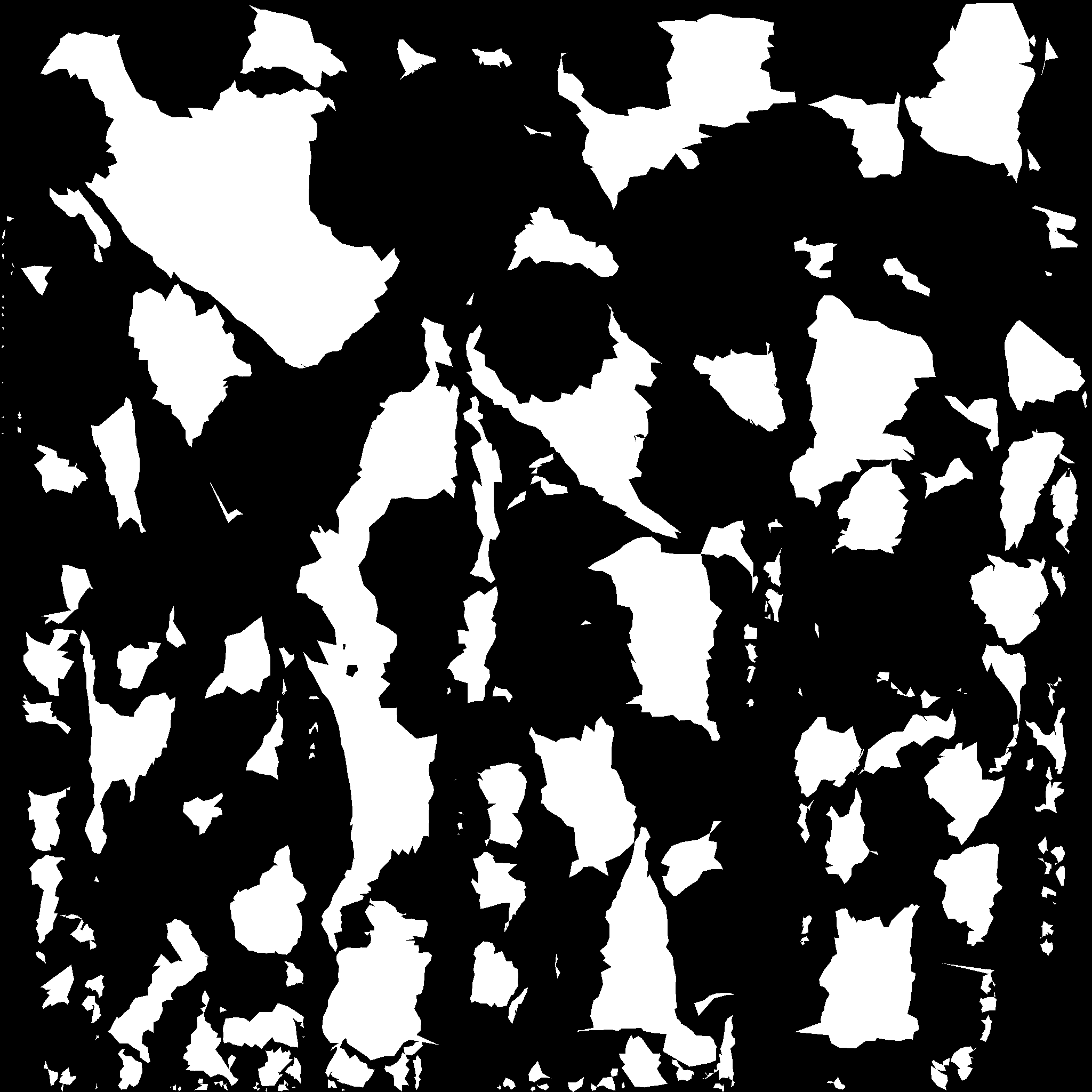}
  \caption{Background mask of partial texture.}
  \label{fig:texture_atlas_partial_mask_background}
\end{subfigure}
\caption{
  Example of complete and partial texture atlases with their corresponding background masks $M_b$.
  Sample from the \bodytextwo{} dataset.
  }
\label{fig:texture_atlas}
\end{figure}

\paragraph{Texture atlas}
A texture image $\mathit{T}$ has a texture atlas structure~\cite{levy2002least} consisting of a
set of charts (\ie small pieces of the body texture) gathered together in a single image.
Each of these charts is mapped onto a different region of the 3D mesh $\mathcal{S}$
using the 2D parametrisation $\pi$.
This allows for densely colouring a 3D mesh from a 2D image.
Fig.~\ref{fig:texture_atlas_complete} shows an example texture atlas corresponding to a complete body mesh.
Fig.~\ref{fig:texture_atlas_complete_mask_background} shows its corresponding background mask.
The background corresponds to the regions outside of the charts.
They do not contain any texture information and are coloured black by convention.
The background is defined in a \emph{background mask} $M_b$
where $M_b(i,j) = 0$ if $(i,j)$ is a background pixel,
and $M_b(i,j) = 1$ otherwise (foreground).
Fig.~\ref{fig:texture_atlas_partial}~and~\ref{fig:texture_atlas_partial_mask_background},
show the partial texture atlas and background mask of a corresponding partial mesh (generated synthetically).

%% file: ProposedApproach.tex
\section{Proposed Approach}
\label{sect:approach}

\begin{figure}
    \centering
    \includegraphics{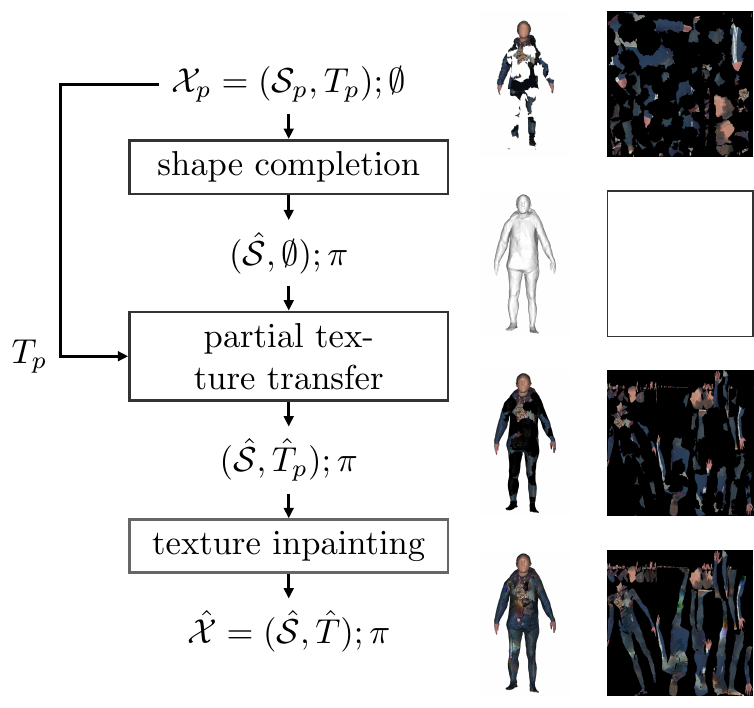}
    \caption{Overview of the proposed approach for 3D body shape and texture completion.}
    \label{fig:approach}
\end{figure}

We propose to solve the problem of textured 3D body shape completion with two sequential tasks:
shape completion, followed by texture completion (Fig.~\ref{fig:approach}).
First, a complete 3D shape $\hat{\mathcal{S}}$ is predicted
from the partial shape $\mathcal{S}_p$ by an encoder-decoder model.
The encoder-decoder completes the input partial shape by deforming a template mesh of a full 3D body
into the corresponding pose and shape of the input.
The texture information of the partial input mesh is then projected
onto the estimated shape, $\hat{\mathcal{S}}$,
to obtain a completed shape with partial texture, $\hat{\mathit{T}}_p$.
The regions with missing texture information are then identified on the texture image, $\hat{\mathit{T}}_p$.
Given the partial texture and the missing regions, the task of texture completion on a 3D shape
is turned into an image inpainting task with additional constraints to handle the specific
image representation of texture atlas.
Indeed, the texture image contains irregularly-shape background regions that must be correctly
ignored to avoid their propagation and unrealistic inpainting results.
The different stages of the approach are detailed below. 

\subsection{3D Body Shape Completion}
\label{sect:shape_comp}

\begin{figure}
    \centering
    \includegraphics{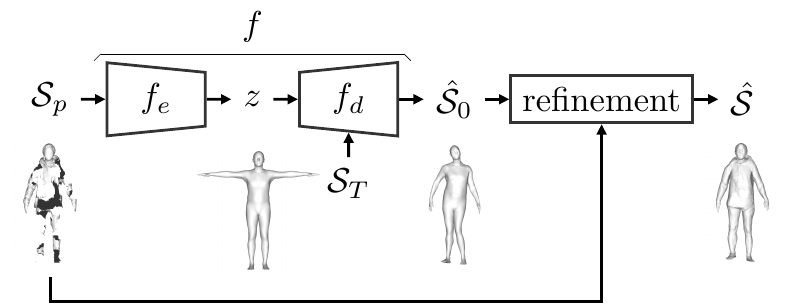}
    \caption{
    Pipeline for 3D shape completion:
    1) An encoder-decoder network produces a first estimate of a complete shape.
    3) The estimate is refined to better fit the clothing.
    }
    \label{fig:approach_shape}
\end{figure}

The 3D body shape completion is performed in two steps, as illustrated in Fig.~\ref{fig:approach_shape}.
First, an encoder-decoder network predicts a rough estimation of the pose and shape of the partial
input shape.
Second, the estimation is refined to better match the clothing shape.

The \emph{encoder-decoder}, $f$, maps a partial 3D shape, $\mathcal{S}_p$, onto a completed shape,
$\hat{\mathcal{S}}$.
This model is based on~\cite{groueix20183d}.
The encoder, $f_e$, transforms the input partial shape into a latent representation, $z\in\mathbb{R}^{n_z}$,
of the body pose and human shape.
The decoder, $f_d$, uses this latent code to deform a template body mesh, $\mathcal{S}_T$,
into the pose and shape of the input.
The result is a first estimation, $\hat{\mathcal{S}}_0$, of the ground-truth complete shape, $\mathcal{S}$.

To refine the first shape estimate, $\hat{\mathcal{S}}_0$, the corresponding latent code $\hat{z}_0$ is
adjusted such that the decoded shape better fits the input partial shape.
This is cast as the optimisation problem
\begin{equation}
\hat{\mathcal{S}}
= \argmin_z d_{\text{Chamfer}}(\mathcal{S}_p, f_d(z))
.
\label{equ:shape_refinement_optim_problem}
\end{equation}
The decoder, $f_d$, is taken as a black-box function.
The objective function is the directed Chamfer distance~\cite{fan2017point} from the partial shape,
$\mathcal{S}_p$,
to the estimation,
$\hat{\mathcal{S}}$.
The directed Chamfer distance is important to only fit the partial shape where there is information
and prevent uncontrolled deformations in holes
(as reported in the experiments in Section~\ref{sect:exp_shape}).
Additionally, the fitting makes use a higher-resolution shape than the one used in the encoder-decoder
to fit clothing shape details more precisely.

\subsubsection{Architecture}
The encoder-decoder is parametrised by a deep neural network with 3D convolutions.
The encoder, $f_e$, follows a PointNet~\cite{qi2017pointnet} architecture.
It takes as input the set of mesh vertices, $\mathbf{V}\in\mathbb{R}^{n_v \times 3}$,
and applies successive point convolutions.
A point convolution consists in a shared multi-layer perceptron (MLP) applied
on the features of a set of points, to produce transformed features of a possibly
different size.
The feature size of the successive layers are $(3, 64, 128, 1024)$.
The last layer is max-pooled across the points into a vector, $z_0$,
of size $n_z=1024$.
This vector is then refined using two densely connected layers of size $1024$ to produce
the latent vector $z$ of size $n_z$ as well.
The decoder, $f_d$, concatenates the latent code, $z$, to each vertex of the template
mesh and then applies a series of point convolutions, as above.
The feature size of the layers in the decoder are
$(3 + 1024, 513, 256, 128, 3)$.

\subsubsection{Training}
The training strategy of the network follows~\cite{groueix20183d}.
The mesh of the SMPL body model~\cite{loper2015smpl} is used as the body mesh template, $\mathcal{S}_T$.
The encoder-decoder network, $f$, is trained with supervision
by learning to reconstruct the SMPL body model in randomly generated poses and shapes.
The input training data is augmented with random subsampling of the points and
random shifts in the positions.
This makes the network robust to partial and irregular sampling and
variations in the input data.
The loss is the mean-squared error (MSE) on the point positions.

\subsection{Body Texture Completion}
\label{sect:texture_comp}

After estimating the complete 3D shape (Section~\ref{sect:shape_comp}),
a corresponding complete texture image is estimated with the following steps.
First, the input partial texture is transferred onto the texture image of
the estimated shape.
Then, the regions to be inpainted are identified.
Finally, the completion of the texture is performed by inpainting the partial
texture image with specific constraints for handling the topology of the texture atlas.
These steps are detailed below.

\subsubsection{Partial texture transfer}
The method of Section~\ref{sect:shape_comp}
estimates a complete mesh $\hat{\mathcal{S}}$
aligned with the input partial shape $\hat{\mathcal{S}}_p$.
The partial texture $\mathit{T}_p$ of the input mesh
is transferred onto $\hat{\mathcal{S}}$
by a ray-casting algorithm that propagates the texture information
along the normal directions.
The result is a mesh with a complete shape and a partial texture, $\hat{\mathcal{X}}_{T_p} = (\hat{\mathcal{S}},\hat{\mathit{T}_p})$.
This is illustrated by the mesh in Fig.~\ref{fig:partialtex_on_comp},
where the regions without mapped colour are rendered in black
(default background colour).
These regions must be identified on the corresponding texture image prior to inpainting.

\subsubsection{Identification of the regions with missing texture}
\begin{figure}
\centering
\begin{subfigure}{.2\textwidth}
  \centering
  \includegraphics[height=\linewidth]{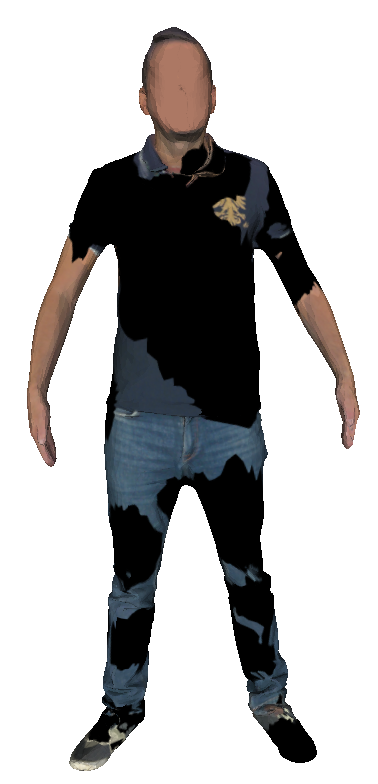}
  \caption{Partial texture transferred on a complete mesh.}
  \label{fig:partialtex_on_comp}
\end{subfigure}%
\hfill
\begin{subfigure}{.2\textwidth}
  \centering
  \includegraphics[height=\linewidth]{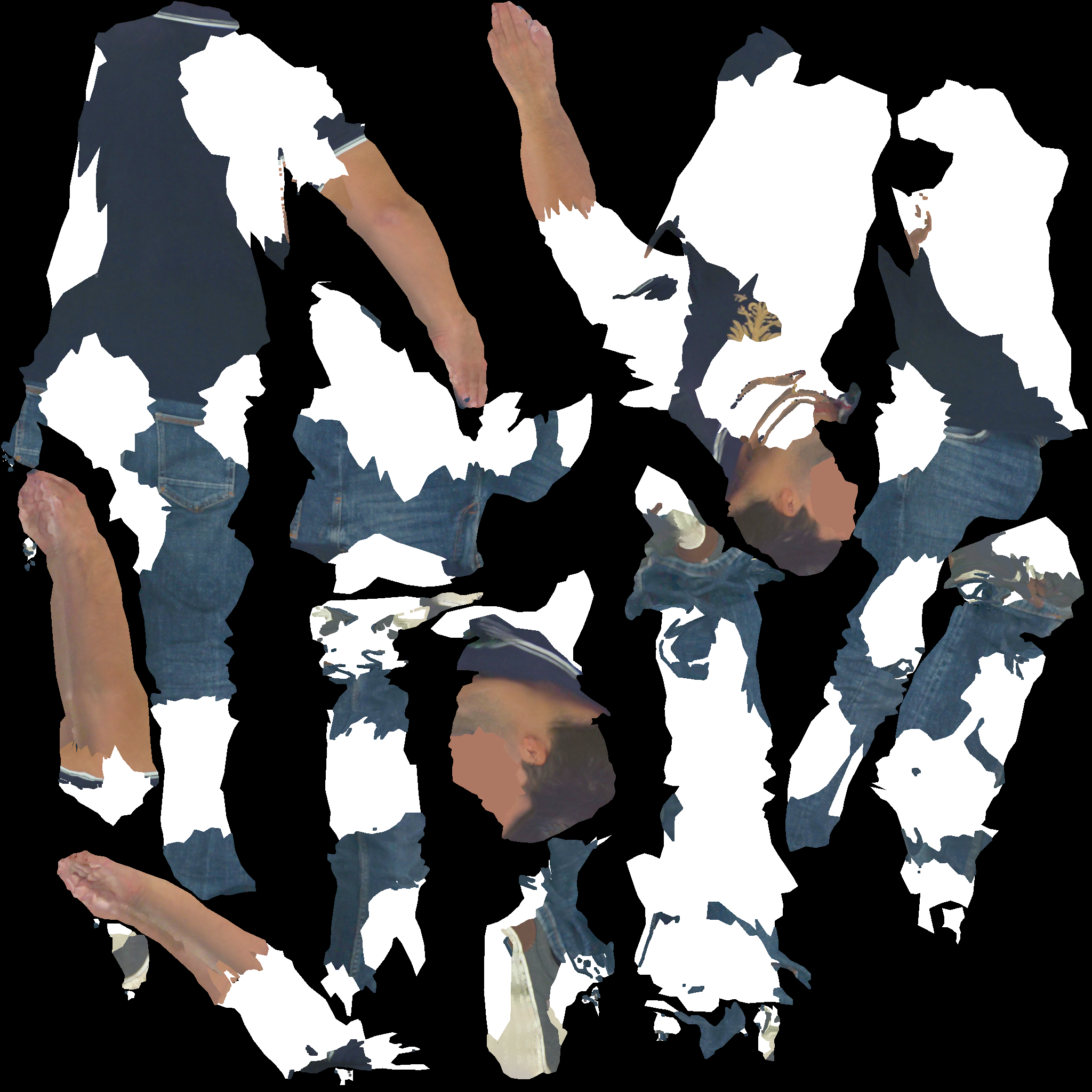}
  \caption{Identified regions with missing texture (white).}
  \label{fig:identified_missing_partial}
\end{subfigure}
\hfill
\begin{subfigure}{.2\textwidth}
  \centering
  \includegraphics[height=\linewidth]{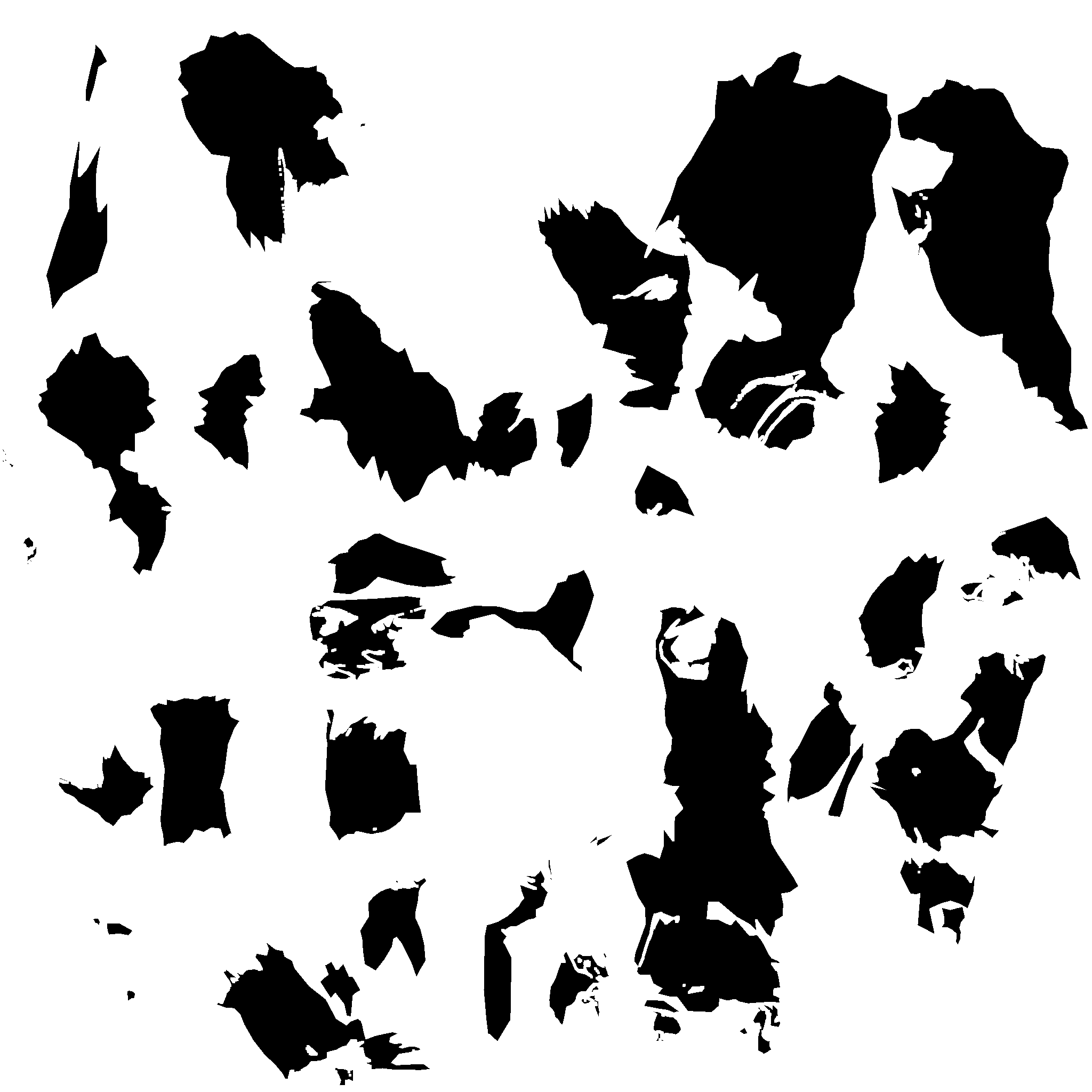}
  \caption{
  Binary mask of the partial texture.
  \\ 
  }
  \label{fig:binary_mask_partialtex}
\end{subfigure}
\hfill
\begin{subfigure}{.2\textwidth}
  \centering
  \includegraphics[height=\linewidth]{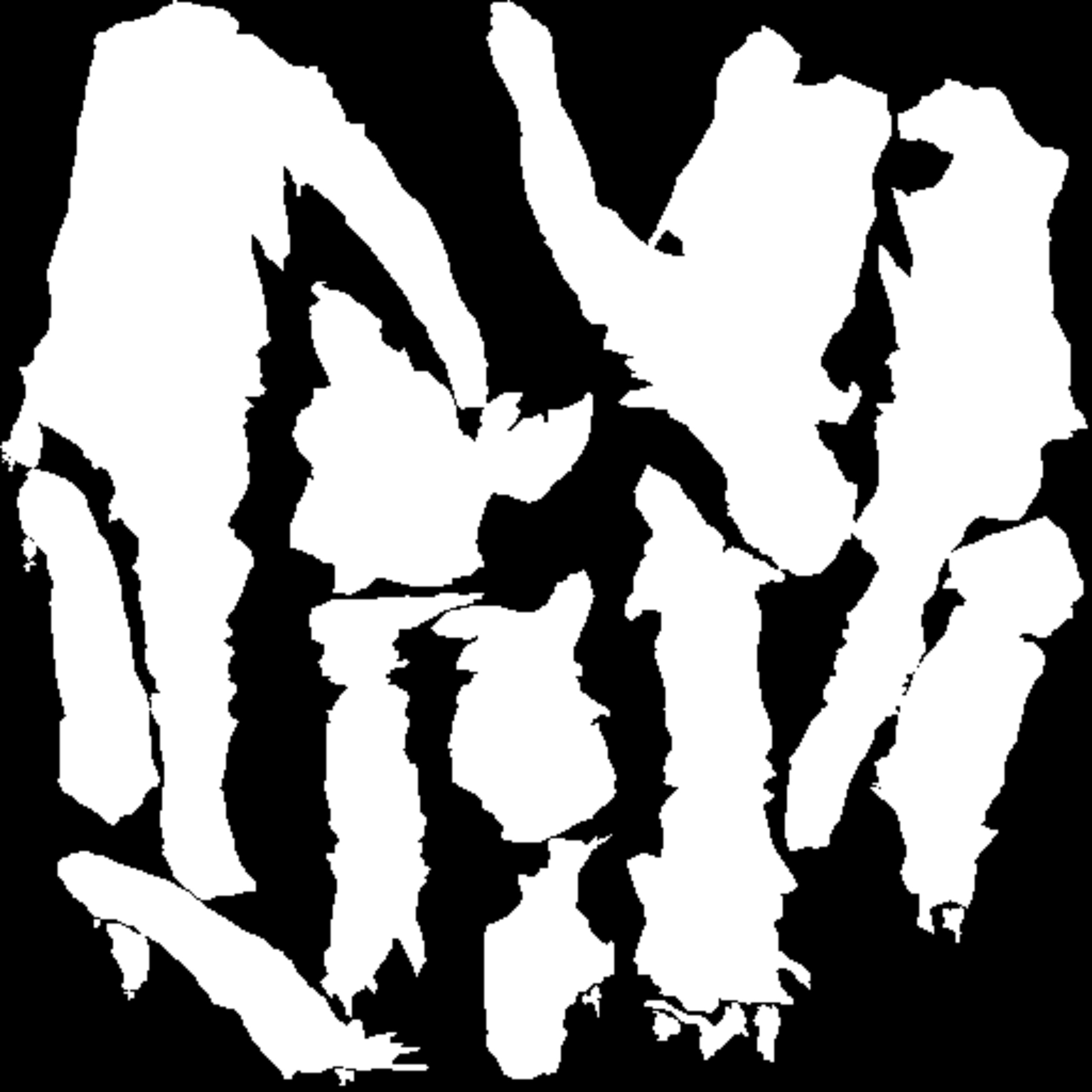}
  \caption{
  Binary background mask.
  \\ 
  \\ 
  }
  \label{fig:binary_mask_backgournd}
\end{subfigure}
\caption{Identification of missing regions on a partial texture and binary masks calculation.}
\label{fig:id_partialtex}
\end{figure}

If no texture information is transferred in a particular region of the mesh
$\hat{\mathcal{X}}_{T_p}$,
the corresponding region in the texture image is left unmodified with the
default black background colour.
Thus, the black pixels inside the charts of the texture atlas
indicate missing texture information.
This is illustrated in Fig.~\ref{fig:identified_missing_partial}
where the identified regions without texture are highlighted in white.
A binary mask $M$ with the same dimension as the partial texture image
is derived such that $M(i,j)=0$
if the pixel $(i,j)$ corresponds to a missing texture information
due to partial data,
and $M(i,j)=1$ otherwise.
An example of the computed mask is shown in Fig.~\ref{fig:binary_mask_partialtex}. The corresponding background mask $M_b$ is shown in Fig.~\ref{fig:binary_mask_backgournd}.


\subsubsection{Texture inpainting}
To the transferred partial texture image $\hat{\mathit{T}_p}$
is associated the mask $M$ indicating the regions with missing information.
Additionally, the background mask $M_b$ of the texture is known from the definition
of the charts in the texture atlas (Section~\ref{sect:problem_statement}).
Given this information, the problem of texture completion is turned into an image inpainting task.
However, the inpainting should only occur in the foreground regions, \ie on the charts of the texture atlas.
Moreover, the non-informative background of the image must be explicitly ignored
to prevent irrelevant background colour (\eg black) to propagate onto the charts.
The proposed adapted inpainting algorithm is detailed below.

Image inpainting is extensively studied in the literature~\cite{yeh2017semantic,yu2019free,yu2018generative,liu2018partialinpainting}.
While some works focus on image inpainting with regular masking shapes (\eg rectangular masks)~\cite{yu2018generative,yeh2017semantic},
more recent works try to address the inpainting problem in case of irregular masking shapes~\cite{liu2018partialinpainting,yu2019free}.
In our case, the masks of the missing texture are derived from partial 3D shapes.
This makes the masks irregular and not restricted to specific shapes as it can be observed in Fig.~\ref{fig:binary_mask_partialtex}.
Consequently, the selected image inpainting approach should take into account these irregularities.
Accordingly, we build on the method proposed in~\cite{liu2018partialinpainting} handling irregular masks.
In~\cite{liu2018partialinpainting}, the authors use partial convolutional layers instead of conventional convolutional layers.
These layers consist of mask-aware convolutions and a mask update step. 

Given a binary mask $M$, partial convolutions extend standard convolutions
to focus the computations on the information from unmasked regions
(\ie pixels $(i,j)$ such that $M(i,j) = 1$)
and discard the information from masked regions.
With the goal of inpainting, the masks are updated after every partial convolutional operation by removing the masking
(\ie changing the mask value from $0$ to $1$) for each location that was involved in the convolution. 
The mask update forces the masked regions to disappear after a sufficient number of updates. More formally, let $W$ be  the weights for a specific convolution filter and $b$ its corresponding bias. $T_w$ are the  feature values for the current sliding window and $M$ is the corresponding binary mask. The partial convolution at at every location, similarly defined in~\cite{liu2018partialinpainting}, is expressed as:

\begin{equation}
t_c =
\begin{cases}
  W^T (T_w \odot M)
  \cdot
  \frac{\text{sum}(\mathbf{1})}{\text{sum}(M)} + b
  &
  \text{if}\; \text{sum}(M) > 0
  \\
  0
  &
  \text{otherwise}
  \end{cases}
\end{equation}

\noindent where $\odot$ denotes element-wise multiplication, and $\mathbf{1}$ has same shape as $M$ but with  all  elements  being  1. After every partial convolution, a masked value in $M$ ($M(i,j)=0$) is updated to unmasked ($M(i,j)=1$) if the convolution was able to condition its output on at least one valid input value. In practice this is achieved by applying fixed convolutions, with the same kernel size as the partial convolution operation, but with weights identically set to 1 and no bias. 

One important observation in the two texture atlases provided in Fig.~\ref{fig:texture_atlas}, is that they contain some non-informative black regions used as background to gather the body charts in a single image.
The inpainting of the missing texture information (white regions in Fig.~\ref{fig:identified_missing_partial}) could be impacted by the non-informative background (\ie black) using the original form of partial convolutions introduced in~\cite{liu2018partialinpainting}.
This is confirmed and visualised by experiments in Section~\ref{sect:exp_inpaint}.
As a solution, we propose to ignore these regions during the partial convolutions as done with the masked values of the missing texture to be recovered.
However, these regions should not be updated during the mask update as the background mask should stay fixed
through all the partial convolution layers.
This is achieved by including the \emph{background mask} $M_b$ of the texture image
in the partial convolution as follows, 

\begin{equation}
t_c =
\begin{cases}
  W^T (T_w \odot M \odot M_b)
  \cdot
  \frac{\text{sum}(\mathbf{1})}{\text{sum}(M \odot M_b)} + b
  & \text{if}\; \text{sum}(M \odot M_b) > 0
  \\
  0
  & \text{otherwise}
\end{cases}
\end{equation}

The background mask $M_b$ is passed to all partial convolutions layers without being updated by applying $\emph{do-nothing}$ convolution kernels with the same shape as the ones used for the masks $M$. A $\emph{do-nothing}$ kernel consists of a kernel with zeros values everywhere except for the central location which is set to 1. Moreover, before updating the original mask $M$ we apply on it this background mask $M_b$ by element-wise multiplication so that we guarantee that the mask $M$ will not be updated using the background regions. 

The aforementioned partial convolutional layers are employed in a UNet-like
architecture~\cite{ronneberger2015u} instead of standard convolutions.
Several loss functions are used to optimise the network.
Two pixel-wise reconstruction losses are defined separately on the masked and unmasked regions with a
focus on masked regions.
Style transfer losses are also considered by constraining the feature maps of the predictions and their auto-correlations to be close those of the ground truth~\cite{gatys2015neural}. Finally a Total-Variation (TV) loss~\cite{johnson2016perceptual} is employed on the masked regions to enforce their smoothness.
For more details about the aforementioned inpainting method are presented in~\cite{liu2018partialinpainting}.

The inpainting task is facilitated by the fact that the texture atlas images to inpaint have a fixed arrangement
of the charts.
This is because they are all defined on the same template mesh $\mathcal{X_T}$
of used for the shape completion in Section~\ref{sect:shape_comp}.
This means that semantic body regions are placed consistently on the texture images,
regularising the inpainting problem.

In the experiments (Section~\ref{sect:exp_inpaint}),
two training strategies are investigated for this inpainting method.
First, the network is trained from scratch on the \bodytextwo{} dataset~\cite{Sharp2020}.
Second, the network, pretrained on the ImageNet dataset~\cite{deng2009imagenet}, is fine-tuned on the \bodytextwo{} dataset~\cite{Sharp2020}.


%% file: Experiments.tex
\section{Experiments}
\label{sect:experiments}

This work focuses on the completion of textured 3D human shapes using the \bodytextwo{} dataset introduced in the SHARP challenge~\cite{Sharp2020}.

The \emph{\textbf{SHA}pe \textbf{R}ecovery from \textbf{P}artial Textured 3D Scans} (SHARP)~\cite{Sharp2020}
challenge aims at advancing the research on the completion of partial textured 3D shape.
Two challenges are proposed with two corresponding datasets of 3D scans:
\bodytextwo{}, a dataset of human scans,
and 3DObjectTex, a dataset of generic objects.
\bodytextwo{} contains about 2500 humans scans of a few hundred people
in varied poses and clothing types.
It is an extension of 3DBodyTex~\cite{saint20183dbodytex}.

\subsection{Shape completion}
\label{sect:exp_shape}

\begin{figure}
    \includegraphics[height=0.15\linewidth]{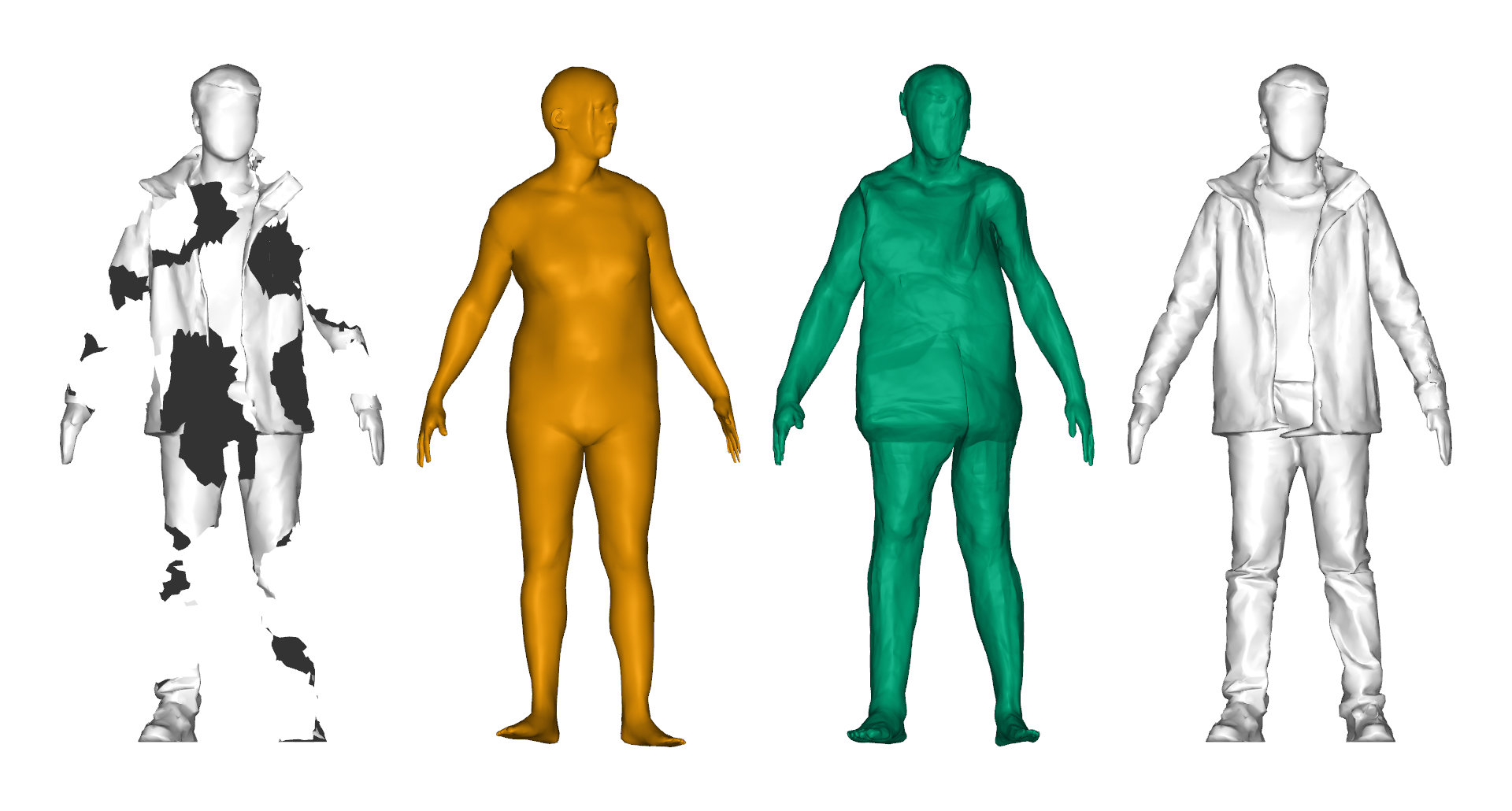}
    \hfill
    \includegraphics[height=0.15\linewidth]{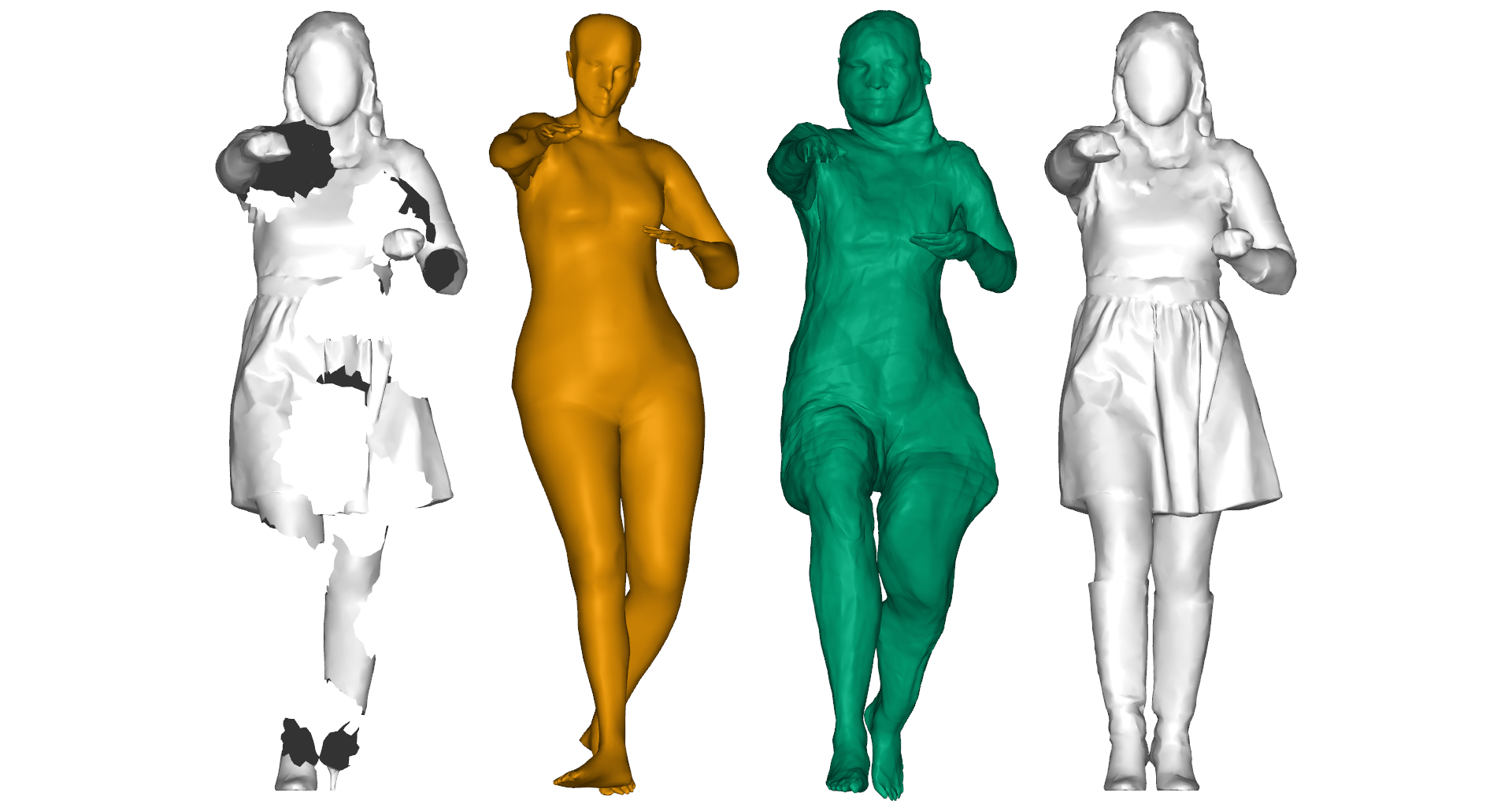}
    \hfill
    \includegraphics[height=0.15\linewidth]{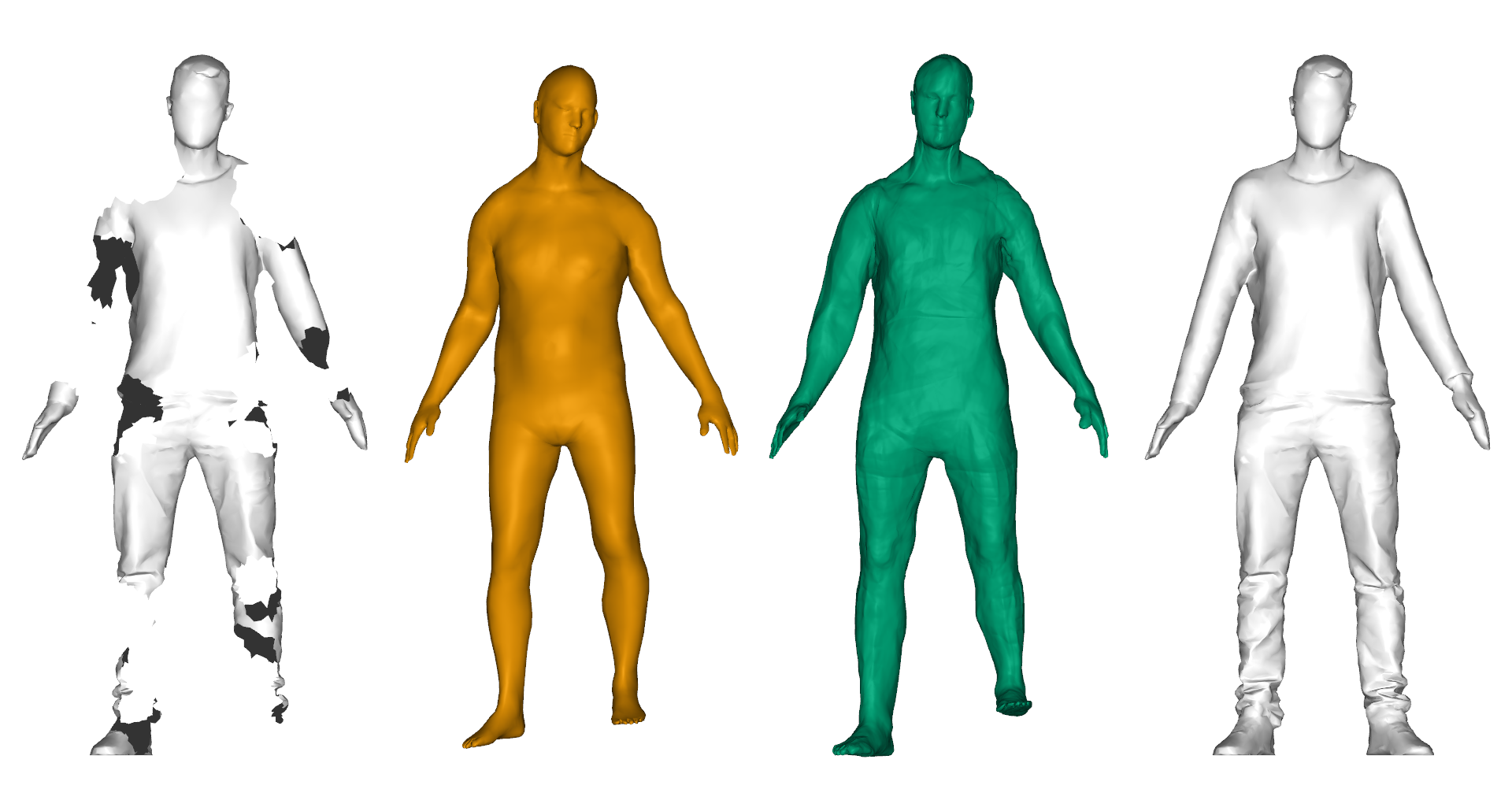}
    \\
    \includegraphics[height=0.15\linewidth]{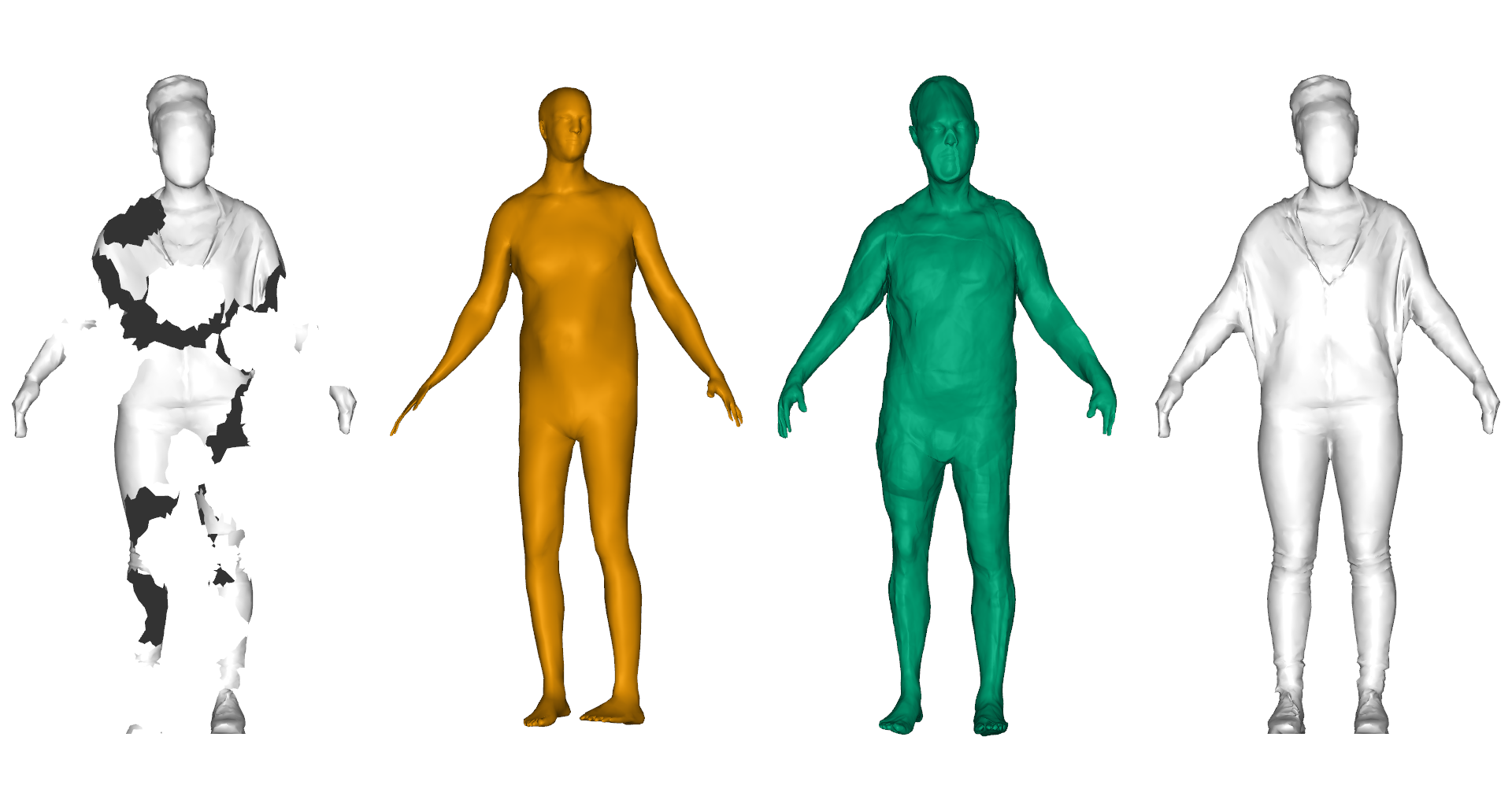}
    \hfill
    \includegraphics[height=0.15\linewidth]{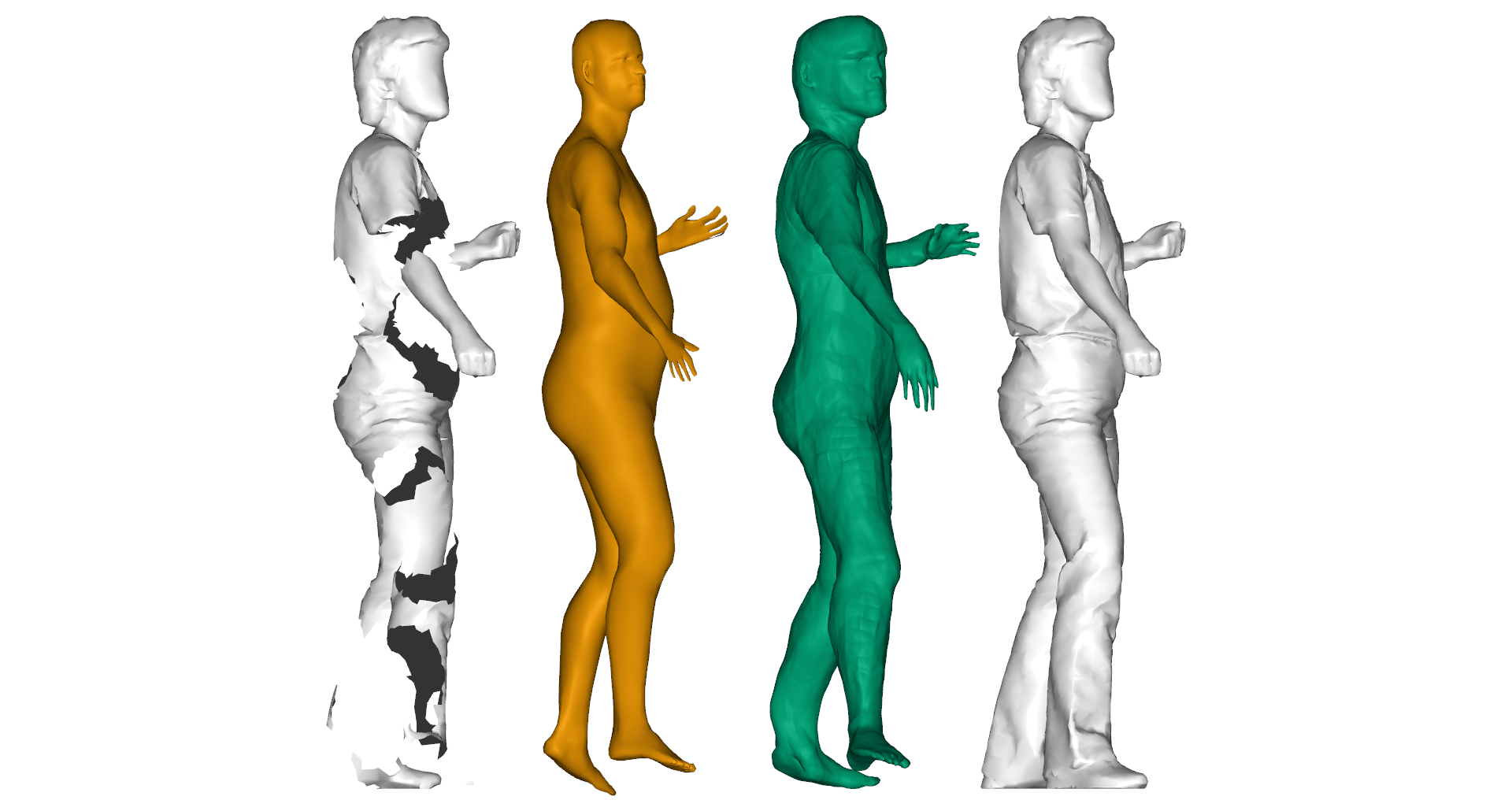}
    \hfill
    \includegraphics[height=0.15\linewidth]{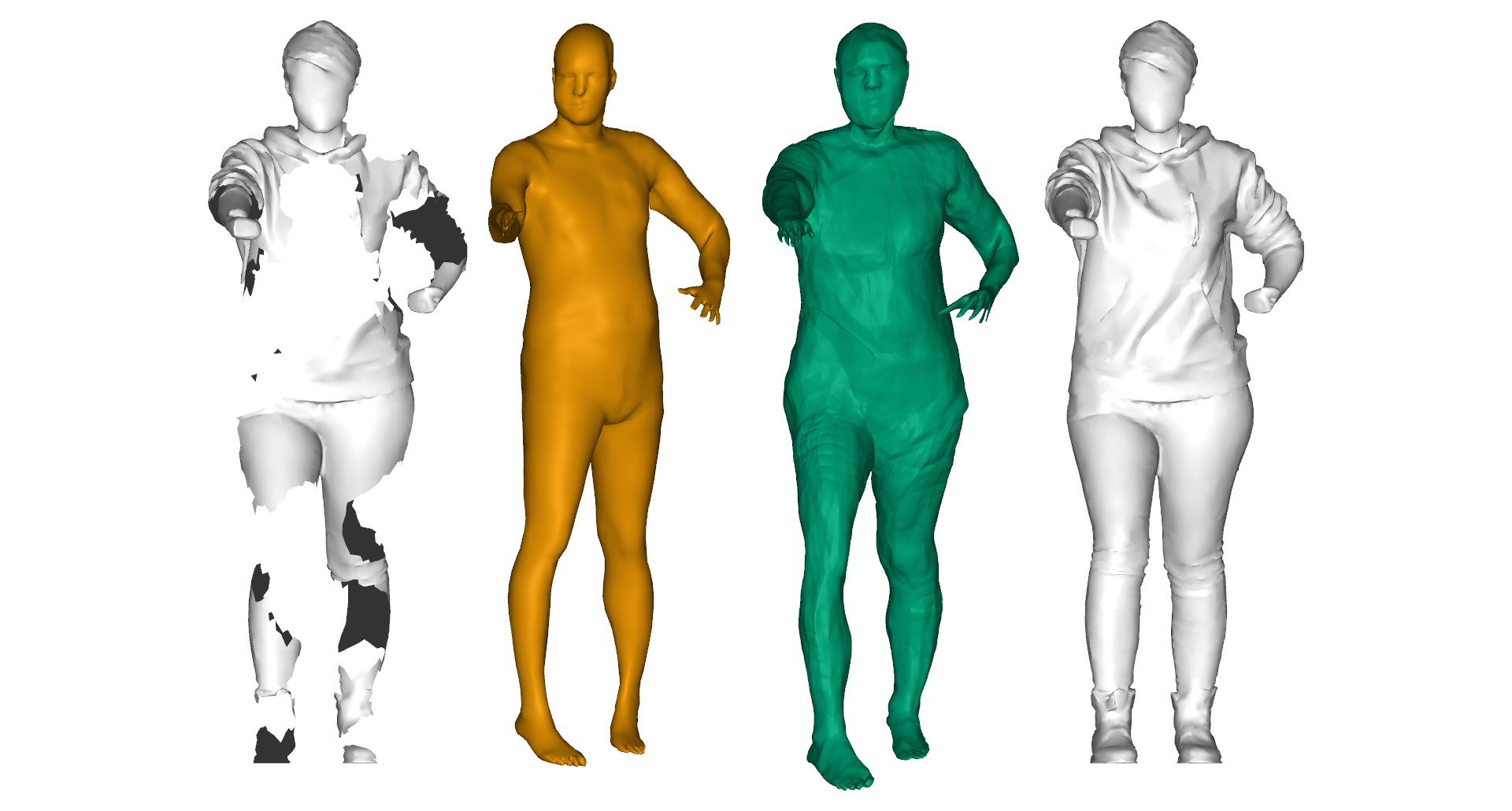}
\caption{
    Results of shape completion for 6 examples of the test set.
    From left to right:
    input partial shape (white),
    initial shape estimate (orange),
    refined shape estimate (green),
    ground truth (white).
    In the input partial shapes, the visible interior surface is rendered in black.
    }
\label{fig:results_shape}
\end{figure}

Fig.~\ref{fig:results_shape} shows the results of the shape completion (Section~\ref{sect:shape_comp}).
It can be seen that initial shape estimate (orange)
captures the pose of the partial input
but not the loose-fitting clothing.
The refined shape (green) represents the clothing more accurately.
This shows the validity of the approach in recovering clothed body shapes.

However, several limitations are observed.
A topology significantly different from the template body mesh is difficulty recovered.
This happens for example with hair and clenched fists.
Moreover, in the example in row 1 column 3 of~Fig.~\ref{fig:results_shape}, the left foot is not 
recovered because it is completely cropped from the partial input.
This suggests that the shape estimation can fail locally on an extremity of the body when no information is
available in the partial shape.
This could be improved by retraining the encoder-decoder model on a dataset of partial shapes.
Furthermore, the pose of the human skeleton is not always sufficiently accurate in the first shape estimate
produced by the encoder-decoder network.
As a consequence, the refinement fails and the final shape estimate is not correctly aligned.
This is due in part to the variety in clothing shapes for which the network has not been trained.
Similarly, in row 1 column 2 of Fig.~\ref{fig:results_shape}, the refined shape (green) does not 
capture the shape of the skirt realistically.
This is also due to the network being trained on body shapes only.
The encoder-decoder, $f$, could thus profit from a training or fine-tuning on a dataset of clothed shapes
and also possibly from adaptations of the architecture to handle the more complex deformations of the clothing.

\begin{figure}
    \includegraphics[height=0.15\linewidth]{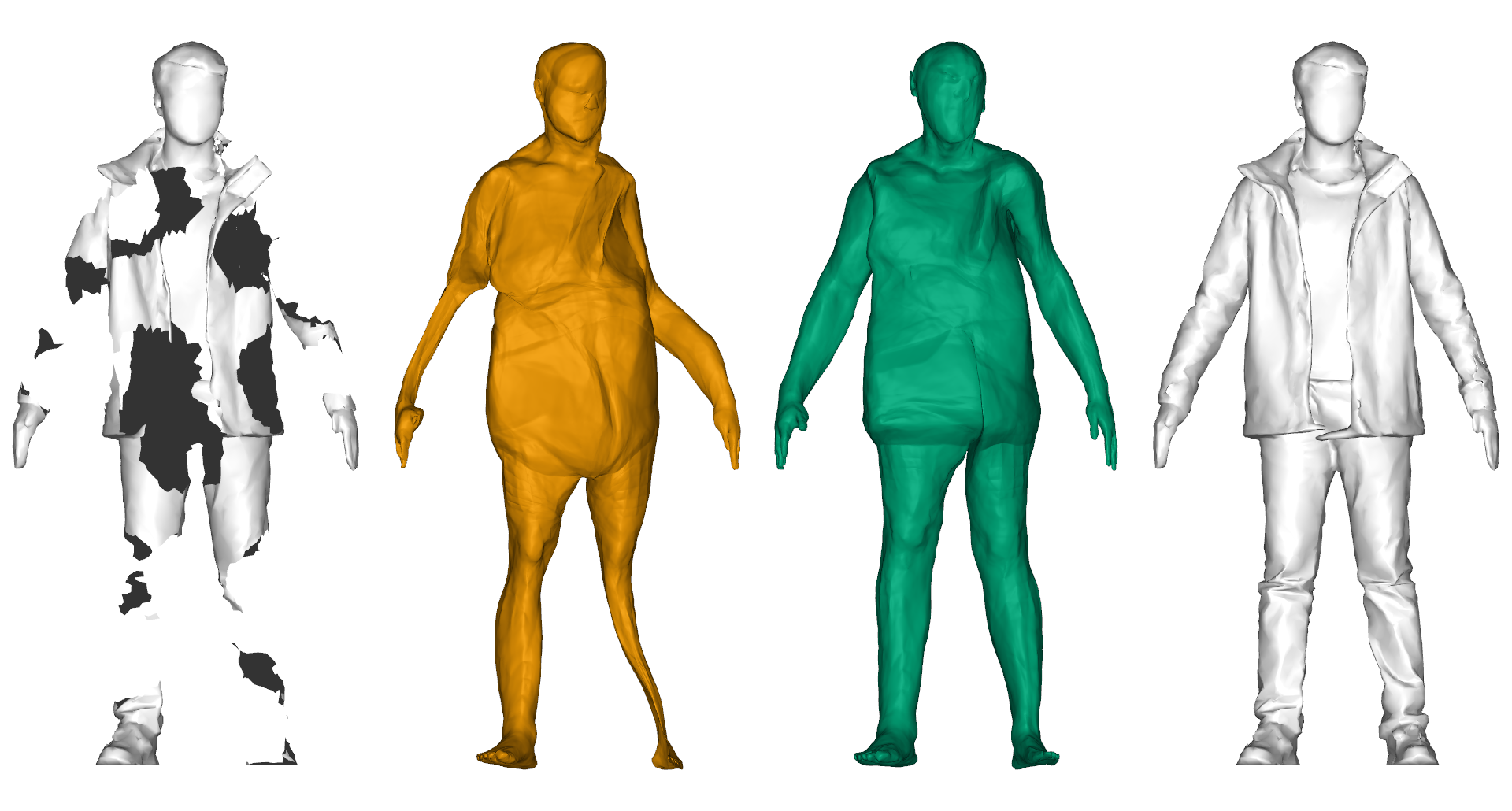}
    \hfill
    \includegraphics[height=0.15\linewidth]{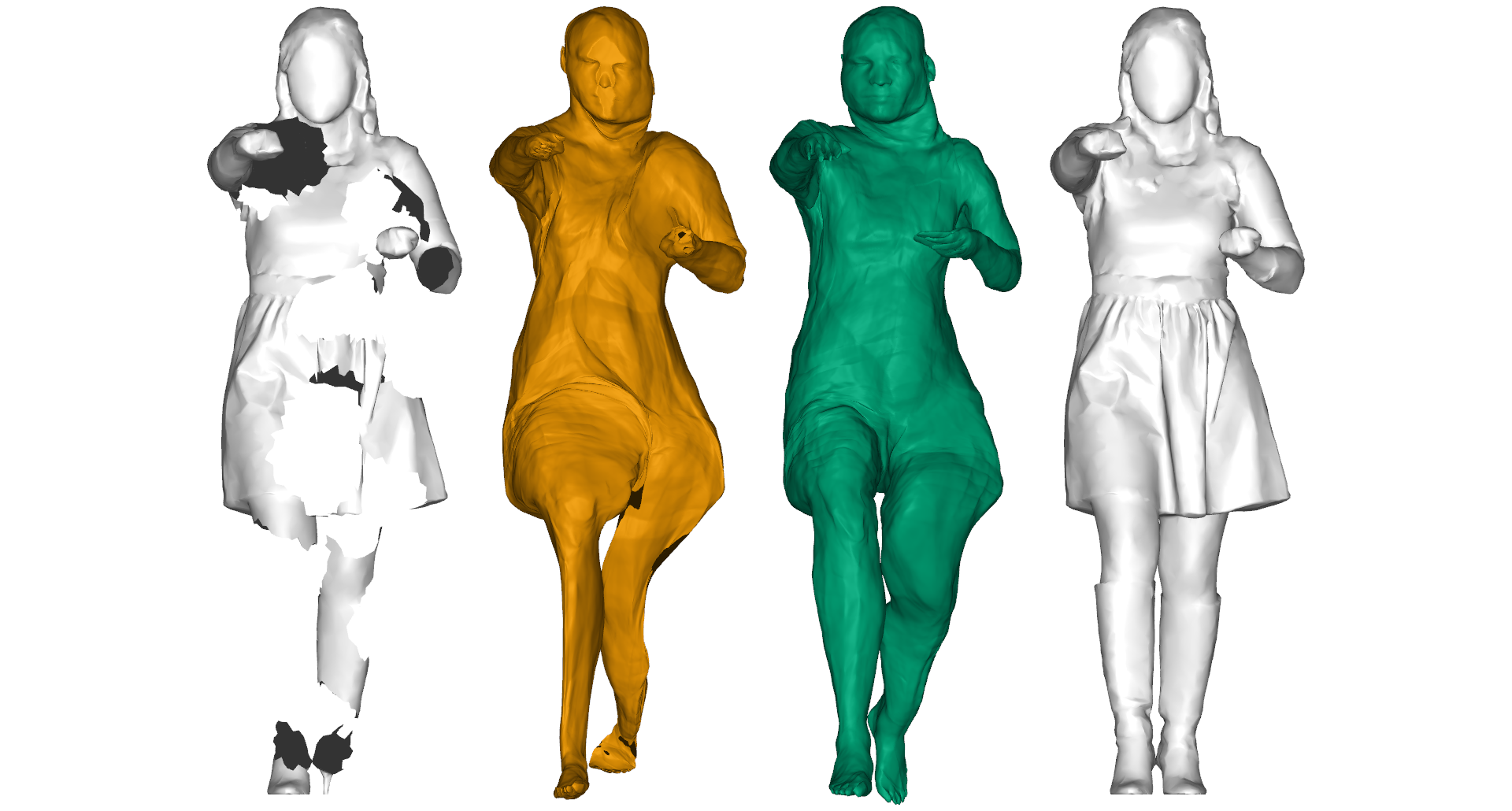}
    \hfill
    \includegraphics[height=0.15\linewidth]{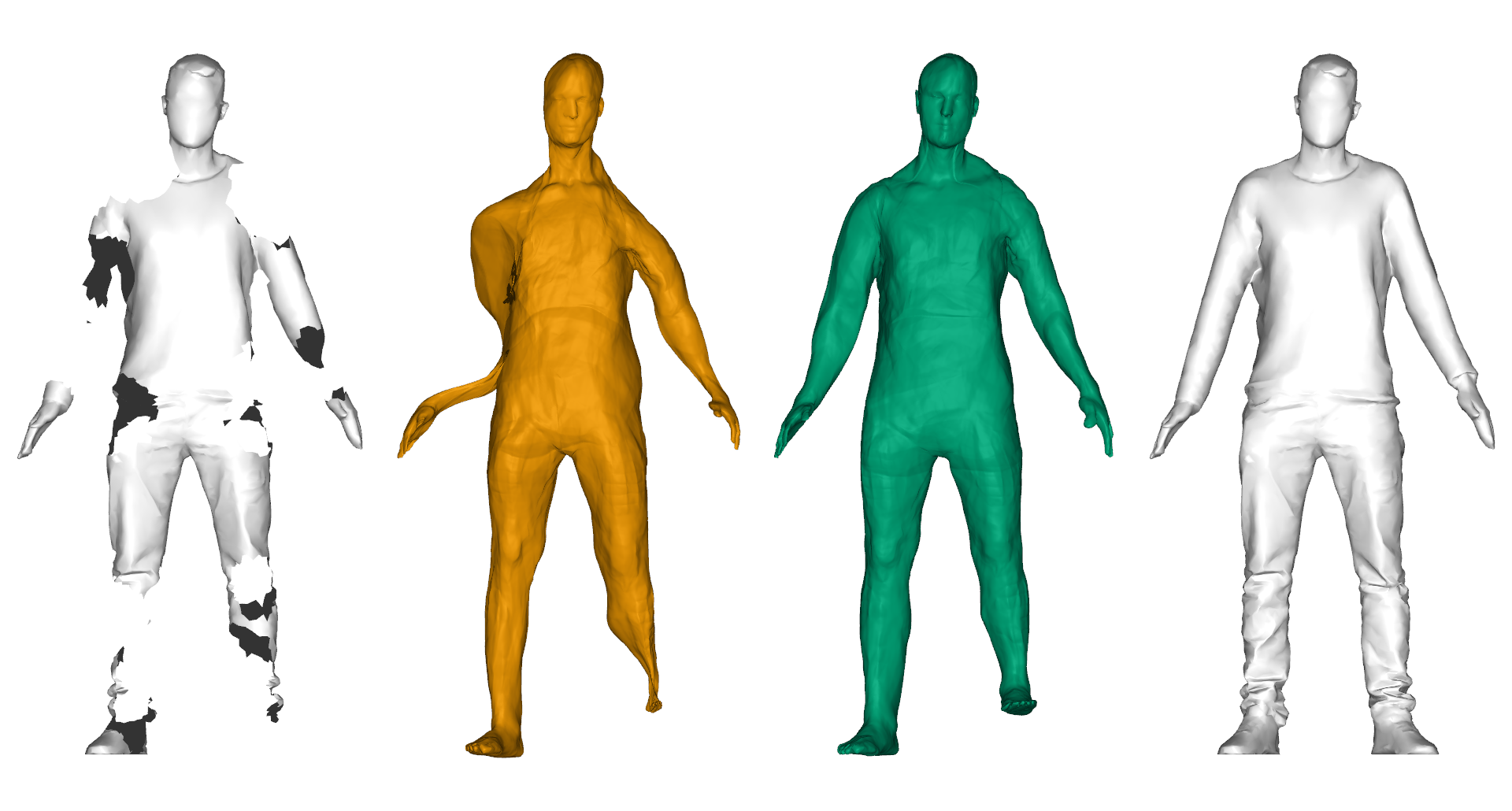}
\caption{
    Shape refinement with symmetric versus directed (one-way) Chamfer distance for three examples of
    the test set.
    The one-way Chamfer distance (green) is the one retained in the proposed approach.
    From left to right:
    input partial shape (white),
    refined shape with symmetric Chamfer distance (orange),
    refined shape with directed (one-way) Chamfer distance (green),
    ground truth (white).
    In the input partial shapes, the visible interior surface is rendered in black.
    }
\label{fig:shape_chamfer12}
\end{figure}

Fig.~\ref{fig:shape_chamfer12} illustrates the importance of the chosen objective function in the
optimisation problem~(\ref{equ:shape_refinement_optim_problem}) to refine the initial shape estimate.
With a symmetric Chamfer distance (from partial input to estimated shape and conversely), the shape
refinement fails (orange shape in~Fig.~\ref{fig:shape_chamfer12}).
With a directed Chamfer distance (from partial input to estimated shape only), the shape refinement is
sound.
This is due to the holes in the input partial shape.
With the symmetric distance, the measure from the estimated shape to the partial input has the effect
of dragging the estimated shape into regions of the partial input without holes.
This creates unrealistic distortions in the estimated shape.

\subsection{Texture transfer}

\begin{figure}
    \includegraphics[height=0.15\linewidth]{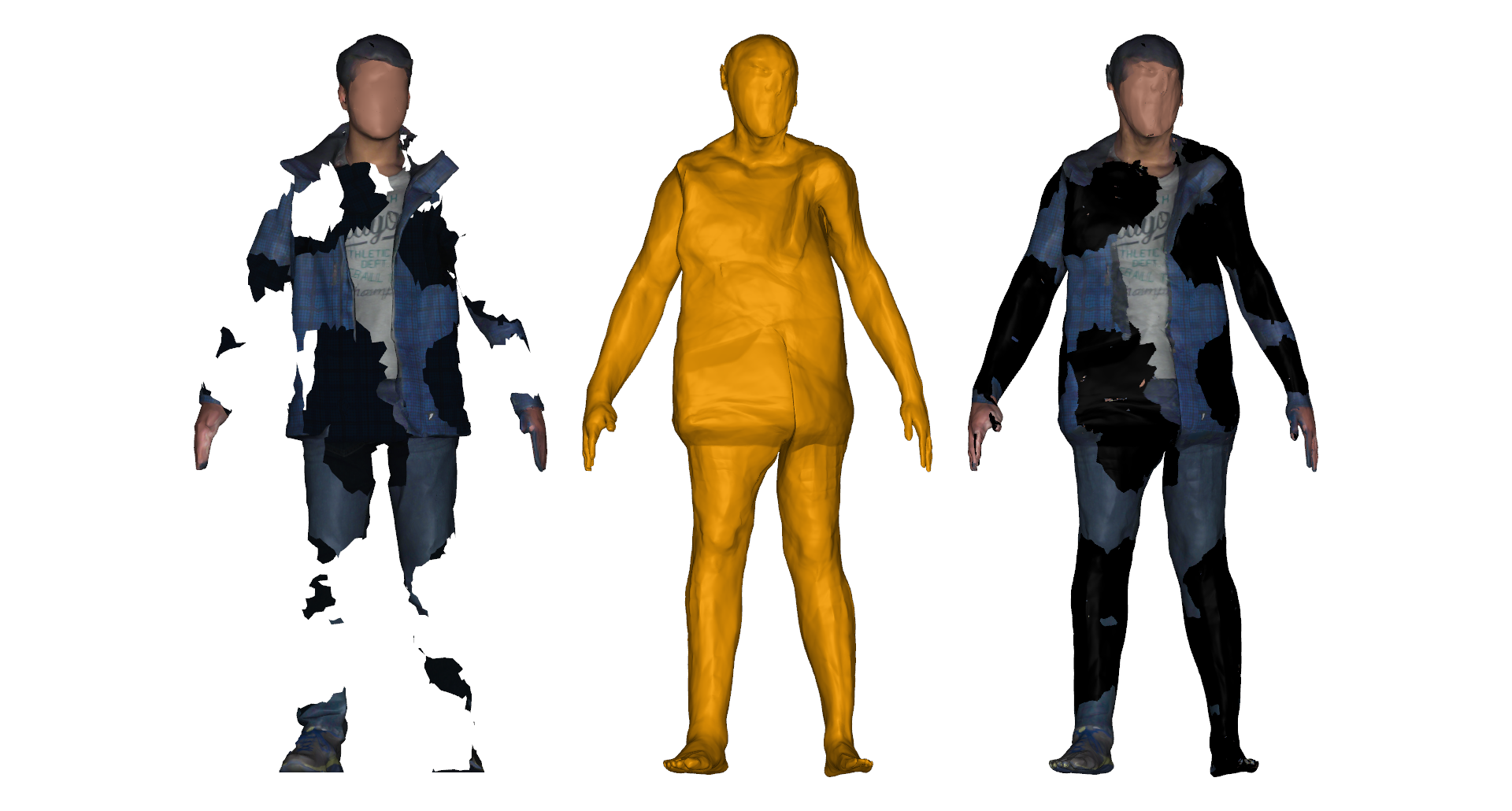}
    \hfill
    \includegraphics[height=0.15\linewidth]{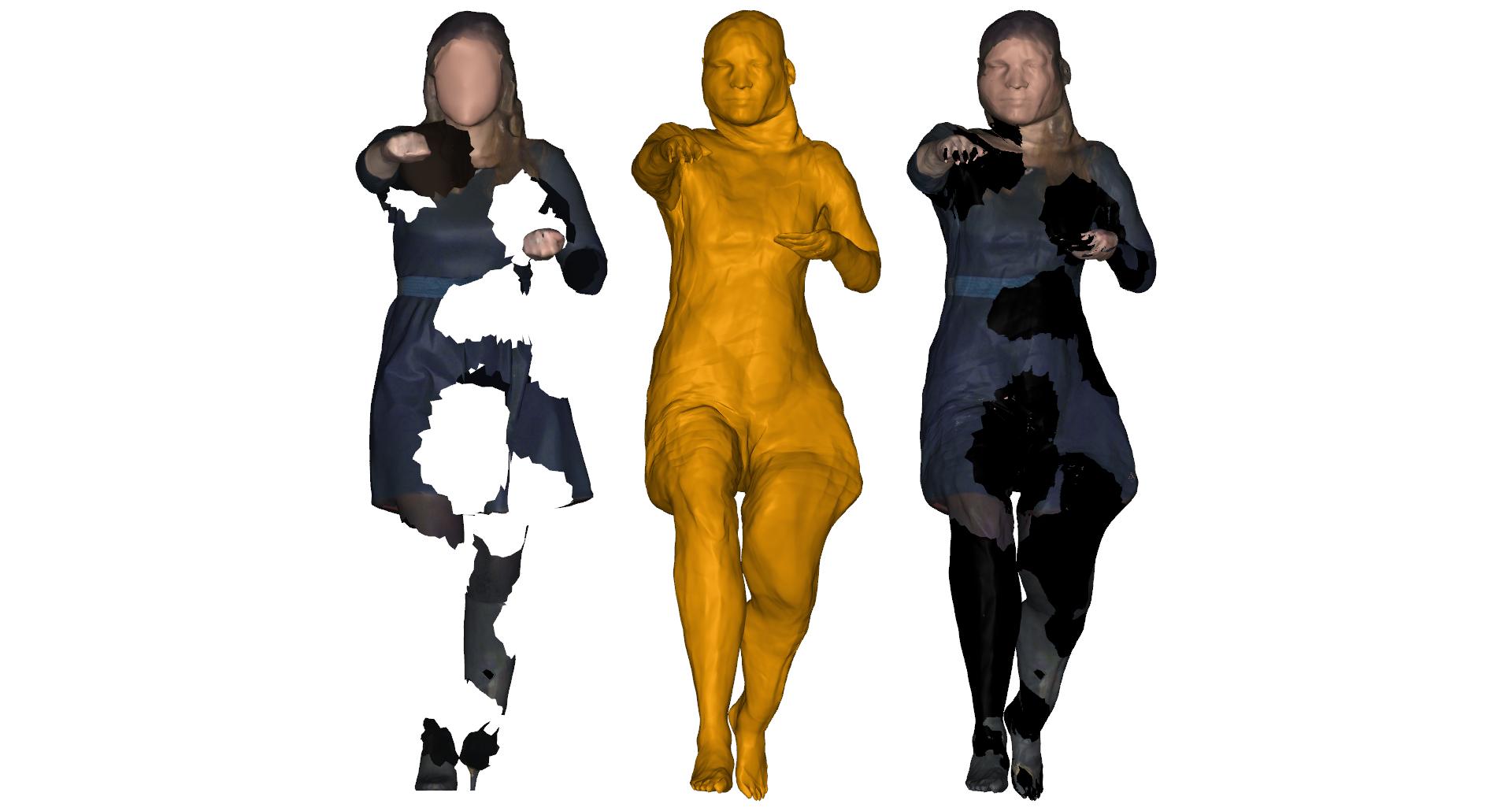}
    \hfill
    \includegraphics[height=0.15\linewidth]{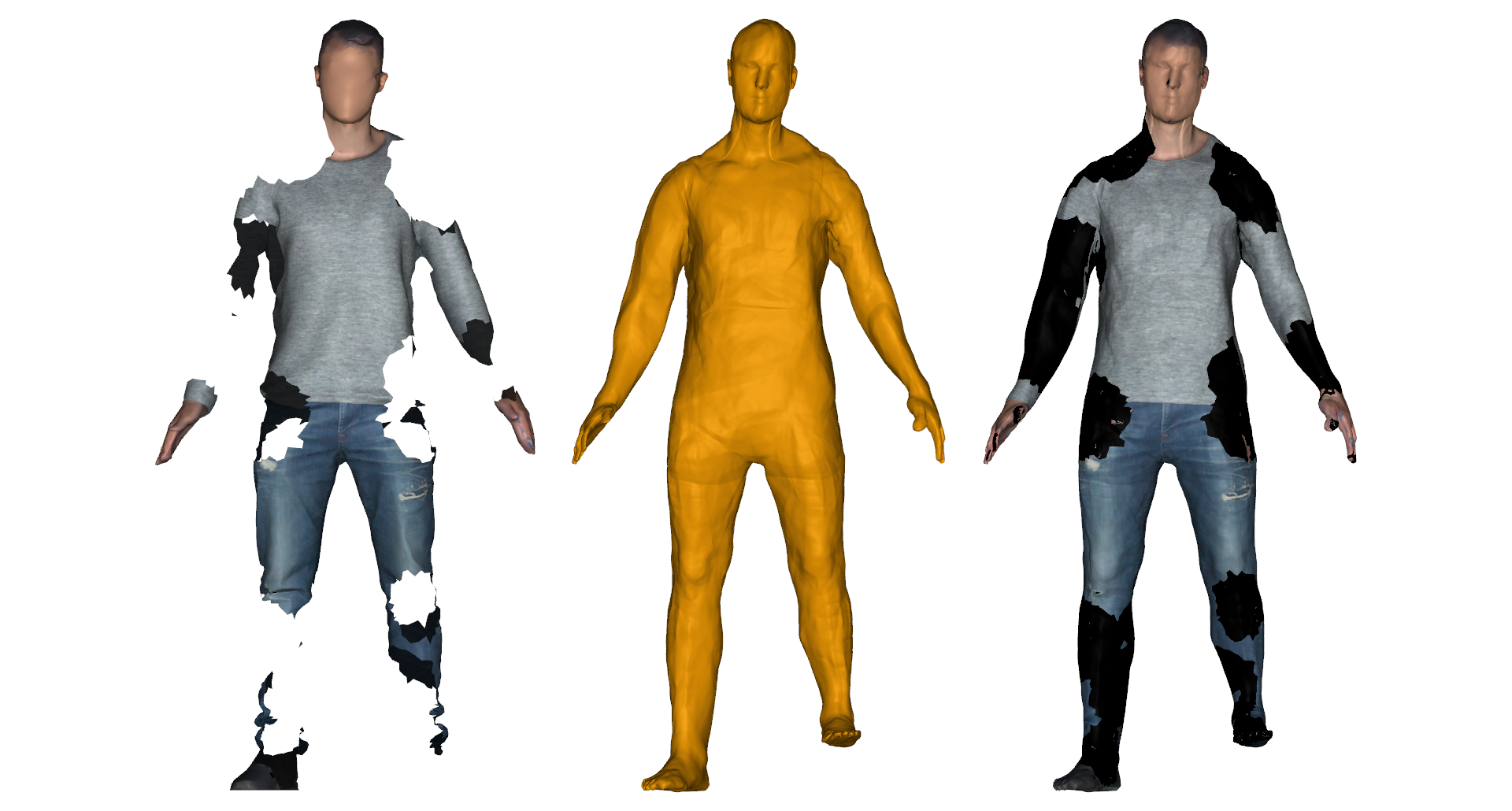}
\caption{
    Illustration of the texture transfer (right)
    from the partial shape (left)
    onto the refined shape estimate (orange, middle)
    for three examples of the test set.
    }
\label{fig:texture_transfer}
\end{figure}

Fig.~\ref{fig:texture_transfer} illustrates the texture transfer from the partial shape onto
the refined shape estimate.
Overall, the texture is mapped correctly when the shape estimate is close the the partial shape.
When the estimated shape is incorrect (e.g. foot in third column), the transferred texture is directly
impacted.

%

\subsection{Texture inpainting}
\label{sect:exp_inpaint}

\begin{figure}

    \centering
    
    \begin{subfigure}{0.15\textwidth}
    \centering
    \includegraphics[height=\linewidth]{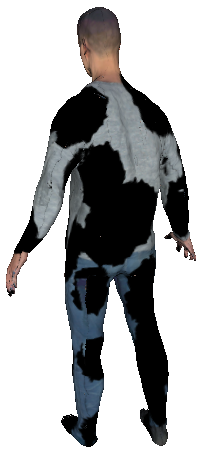}
    \end{subfigure}\hfil
    \begin{subfigure}{0.15\textwidth}
    \centering
    \includegraphics[height=\linewidth]{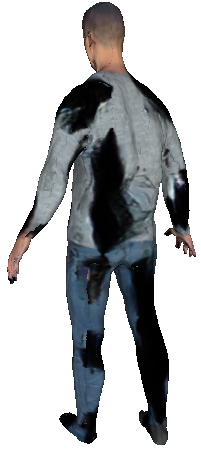}
    \end{subfigure}\hfil
    \begin{subfigure}{0.15\textwidth}
    \centering
    \includegraphics[height=\linewidth]{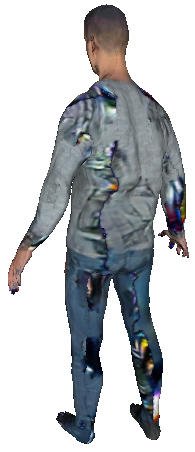}
    \end{subfigure}\hfil
    \begin{subfigure}{0.15\textwidth}
    \centering
    \includegraphics[height=\linewidth]{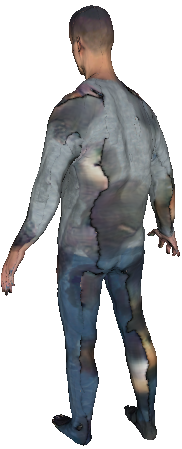}
    \end{subfigure}\hfil
    \begin{subfigure}{0.15\textwidth}
    \centering
    \includegraphics[height=\linewidth]{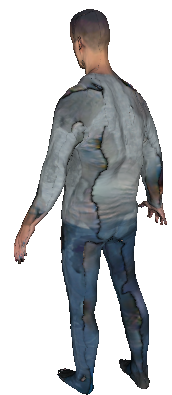}
    \end{subfigure}\hfil
    \begin{subfigure}{0.15\textwidth}
    \centering
    \includegraphics[height=\linewidth]{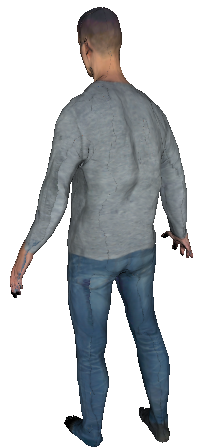}
    \end{subfigure}
    
    \begin{subfigure}{0.15\textwidth}
    \centering
      \includegraphics[height=\linewidth]{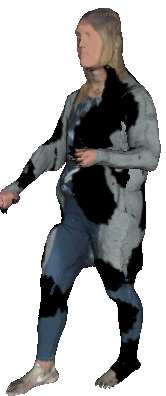}
    \end{subfigure}\hfil
    \begin{subfigure}{0.15\textwidth}
    \centering
      \includegraphics[height=\linewidth]{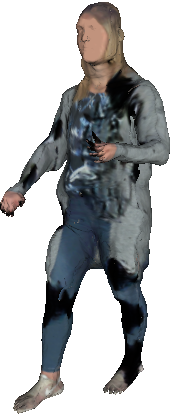}
    \end{subfigure}\hfil
    \begin{subfigure}{0.15\textwidth}
    \centering
      \includegraphics[height=\linewidth]{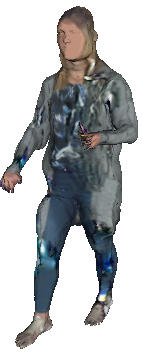}
    \end{subfigure}\hfil
    \begin{subfigure}{0.15\textwidth}
    \centering
      \includegraphics[height=\linewidth]{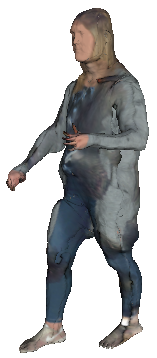}
    \end{subfigure}\hfil
    \begin{subfigure}{0.15\textwidth}
    \centering
      \includegraphics[height=\linewidth]{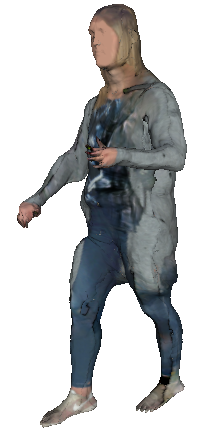}
    \end{subfigure}\hfil
    \begin{subfigure}{0.15\textwidth}
    \centering
      \includegraphics[height=\linewidth]{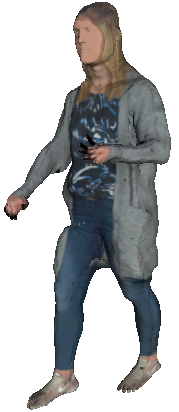}
    \end{subfigure}
    
    \begin{subfigure}{0.15\textwidth}
    \centering
      \includegraphics[height=\linewidth]{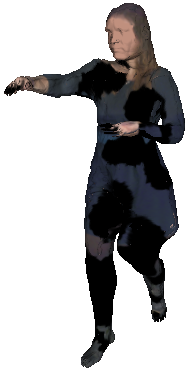}
    \end{subfigure}\hfil
    \begin{subfigure}{0.15\textwidth}
    \centering
      \includegraphics[height=\linewidth]{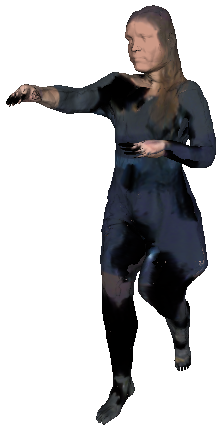}
    \end{subfigure}\hfil
    \begin{subfigure}{0.15\textwidth}
    \centering
      \includegraphics[height=\linewidth]{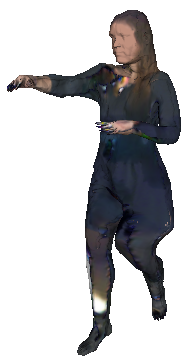}
    \end{subfigure}\hfil
    \begin{subfigure}{0.15\textwidth}
    \centering
      \includegraphics[height=\linewidth]{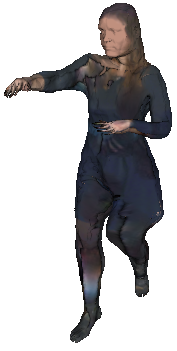}
    \end{subfigure}\hfil
    \begin{subfigure}{0.15\textwidth}
    \centering
      \includegraphics[height=\linewidth]{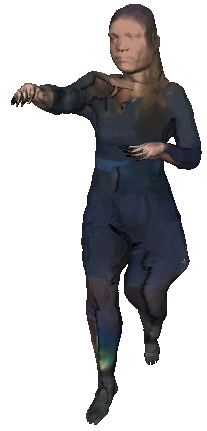}
    \end{subfigure}\hfil
    \begin{subfigure}{0.15\textwidth}
    \centering
      \includegraphics[height=\linewidth]{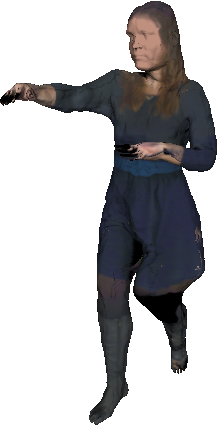}
    \end{subfigure}
    
    \begin{subfigure}{0.15\textwidth}
    \centering
    \captionsetup{singlelinecheck=false,format=hang,justification=justified,font=footnotesize,labelsep=space}
      \includegraphics[height=\linewidth]{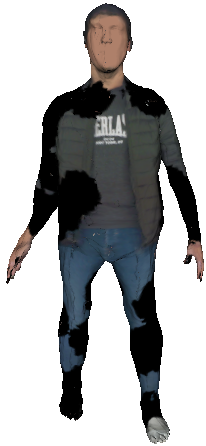}
      \caption{Input partial}
    \end{subfigure}\hfil
    \begin{subfigure}{0.15\textwidth}
    \centering
    \captionsetup{singlelinecheck=false,format=hang,justification=justified,font=footnotesize,labelsep=space}
      \includegraphics[height=\linewidth]{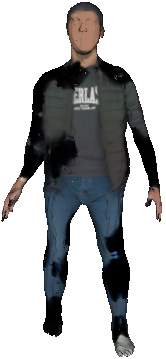}
      \caption{Pretrained\\} 
      \label{fig:inpainting_results_orig_pretrained}
    \end{subfigure}\hfil
    \begin{subfigure}{0.15\textwidth}
    \centering
    \captionsetup{singlelinecheck=false,format=hang,justification=justified,font=footnotesize,labelsep=space}
      \includegraphics[height=\linewidth]{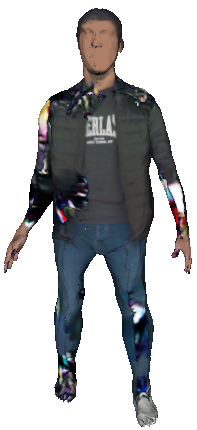}
      \caption{Pretrained + $M_b$}
      \label{fig:inpainting_results_mask_pretrained}
    \end{subfigure}\hfil
    \begin{subfigure}{0.15\textwidth}
    \centering
    \captionsetup{singlelinecheck=false,format=hang,justification=justified,font=footnotesize,labelsep=space}
      \includegraphics[height=\linewidth]{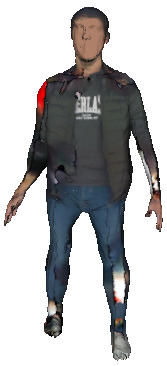}
      \caption{Scratch + $M_b$}
      \label{fig:inpainting_results_mask_scratch}
    \end{subfigure}\hfil
    \begin{subfigure}{0.15\textwidth}
    \centering
    \captionsetup{singlelinecheck=false,format=hang,justification=justified,font=footnotesize,labelsep=space}
      \includegraphics[height=\linewidth]{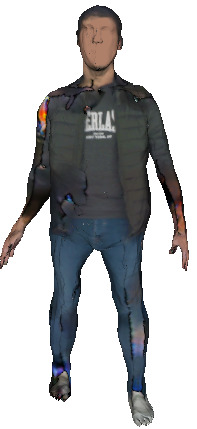}
      \caption{Fine-tuning + $M_b$}
      \label{fig:inpainting_results_mask_finetuned}
    \end{subfigure}\hfil
    \begin{subfigure}{0.15\textwidth}
    \centering
    \captionsetup{singlelinecheck=false,format=hang,justification=justified,font=footnotesize,labelsep=space}
      \includegraphics[height=\linewidth]{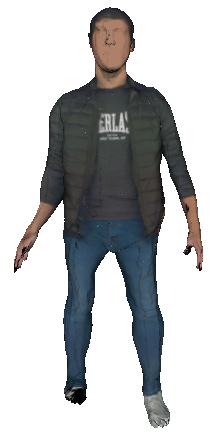}
      \caption{Ground truth }
    \end{subfigure}
    
    \caption{Results for texture inpainting using different strategies.}
    \label{fig:inpainting_results}
\end{figure}

Visual results on the validation set for texture completion are presented in
Fig.~\ref{fig:inpainting_results}.
The completed textures are displayed on the completed shapes obtained with the method from
Section~\ref{sect:exp_shape}.
The first column (a) shows the input partial texture
while the last column (f) depicts the ground-truth complete texture.
The intermediate columns show the inpainting results
and training strategies:
(b) the original inpainting model, pretrained on ImageNet;
(c) the improved inpainting model with the proposed background masks (Section~\ref{sect:texture_comp}),
pretrained on ImageNet;
(d) the improved inpainting model with background masks, trained from scratch on \bodytextwo{}
and (e) the improved inpainting model with background masks,
pretrained on ImageNet
and fine-tuned on \bodytextwo{}.

As seen in Fig.~\ref{fig:inpainting_results_orig_pretrained},
the original inpainting model pretrained on ImageNet
is able to complete some of the missing regions with colour matching the local context.
However, the holes are not completed fully.
This is due to the black background of a texture atlas which is used as a local context for inpainting,
as explained in Section~\ref{sect:texture_comp}.
With the addition of the proposed background masks to the inpainting model
(Fig.~\ref{fig:inpainting_results_mask_pretrained}),
the holes are fully completed.
This validates the proposed approach tailored to the data at hand.
However, the completed colour does not match the local context accurately and contains artefacts.
For example, some holes are filled in with random colour patterns
and some white patches are produced in dark regions.
Training the improved inpainting model from scratch on \bodytextwo{}
(Fig.~\ref{fig:inpainting_results_mask_scratch})
reduces the colour artefacts.
The completed colour patterns are more regular but the colour does not follow closely the
surrounding context.
Thus, the \bodytextwo{} dataset seems enough to regularise the colour pattern inside the clothing
but not rich and varied enough for the model to learn what colour to complete with.
Indeed, \bodytextwo{} is relatively small (a few thousands samples)
in comparison to ImageNet (millions of examples).
Finally, the proposed method of fine-tuning on \bodytextwo{}
the improved inpainting model pretrained on ImageNet
(Fig.~\ref{fig:inpainting_results_mask_finetuned})
gives the best results by producing regular colour patterns matching the local surrounding regions closely.


%% file: Conclusion.tex
\section{Conclusion}
\label{sect:conclusion}

This work proposes \methodname{}, a novel approach for the completion of partial textured human meshes.
The tasks of 3D shape completion and texture completion are addressed sequentially.
First, the 3D body shape completion is performed with an encoder-decoder system
that encodes a partial input point cloud of a human shape into a latent representation of the complete shape
and decodes this representation into a completed shape by deforming a template body mesh.
The estimated shape is further refined to better match the clothing
by adjusting the encoded latent code with an optimisation procedure
adapted to the input partial data.
The partial texture information is then projected onto the estimated shape.
The regions of the mesh to be completed are identified from the 3D shapes and mapped onto the texture image.
The texture completion is then seen as a texture inpainting problem.
For this task, a novel inpainting method tailored to texture maps is proposed.
It is specifically designed
to handle holes of irregular shape
and to be robust to irrelevant background image information.
Experiments on \bodytextwo{} show the validity of the proposed approach on partial data.
Future work includes making the encoder-decoder network capture the clothing shape better,
by training on a dataset of clothing shape and/or adapting the architecture.
The texture transfer could be made more robust to bad shape estimates by enforcing a continuity
in the mapping from partial shape to estimated shape.
Overall, the pipeline might profit from a coupling between the shape and texture completion,
for example in an end-to-end neural network architecture.
This involves at least designing differentiable alternatives to the two intermediate manual steps
of shape refinement and texture transfer.

%% file: acknowledgements.tex
\section*{Acknowledgements}
This work was supported by the Luxembourg National Research Fund (FNR),
projects BODYFIT (11806282)
and IDFORM (11643091),
and Artec3D.